\documentclass[letterpaper, 10 pt, conference]{ieeeconf}  

\IEEEoverridecommandlockouts                              

\makeatletter
\let\NAT@parse\undefined
\makeatother

\usepackage{graphics} 
\usepackage{graphicx}
\usepackage{grffile} 
\usepackage{amsmath} 
\usepackage{amssymb}  
\usepackage{color}
\usepackage{cite}
\usepackage{bm}
\usepackage{url}

\usepackage[noend]{algpseudocode}
\usepackage{algorithm}
\usepackage[table]{xcolor}
\usepackage{makecell}
\usepackage{multirow}
\usepackage[pagebackref=true,breaklinks=true,colorlinks,bookmarks=false]{hyperref}

\def\etal{et al.}

\title{\LARGE \bf
VLASE: Vehicle Localization by Aggregating Semantic Edges
}

\author{\small Xin Yu$^{1*}$, Sagar Chaturvedi$^{1*}$, Chen Feng$^{2}$, Yuichi Taguchi$^{2}$, Teng-Yok Lee$^{2}$, Clinton Fernandes$^{1}$, Srikumar Ramalingam$^{1}$
\thanks{$^{*}$indicate equal contributions.}%
\thanks{$^{1}$University of Utah, Salt Lake City, UT 84112, USA
	{\tt\small \{xin.yu,sagar.chaturvedi,srikumar\}@utah.edu}}%
\thanks{$^{2}$Mitsubishi Electric Research Laboratories (MERL),
	Cambridge, MA 02139, USA
	{\tt\small \{cfeng,taguchi,tlee\}@merl.com}}%
}

\begin{document}

\maketitle
\thispagestyle{empty}
\pagestyle{empty}

\begin{abstract}
In this paper, we propose VLASE, a framework to use semantic edge features from images to achieve on-road localization. Semantic edge features denote edge contours that separate pairs of distinct objects such as building-sky, road-sidewalk, and building-ground. While prior work has shown promising results by utilizing the boundary between prominent classes such as sky and building using skylines, we generalize this approach to consider semantic edge features that arise from 19 different classes. Our localization algorithm is simple, yet very powerful. We extract semantic edge features using a recently introduced CASENet architecture and utilize VLAD framework to perform image retrieval. Our experiments show that we achieve improvement over some of the state-of-the-art localization algorithms such as SIFT-VLAD and its deep variant NetVLAD.
We use ablation study to study the importance of different semantic classes, and show that our unified approach achieves better performance compared to individual prominent features such as skylines.  
\end{abstract}

\section{Introduction}

In the pre-GPS era, we do not describe a location using latitude-longitude coordinates. The typical description of a location is based on certain semantic proximity, such as a tall building, traffic light, or an intersection. While the recent successful image-based localization methods rely on either complex hand-crafted features like SIFT~\cite{lowe2004distinctive} or automatically learnt features using CNNs, we would like to take a step back and ask the following question: How powerful are simple semantic cues for the task of localization? There is a general consensus that the salient features for localization are not always human-understandable, and it is important to capture special visual signatures imperceptible to the eye. Surprisingly, this paper shows that simple human-understandable semantic features, although extracted using CNNs, provide accurate localization in urban scenes and they compare favorably to some of the state-of-the-art localization methods that employ SIFT features in a VLAD~\cite{Jegou12PAMI} framework.

\definecolor{blk_color_0}{rgb}{1.000,0.400,0.000}
\definecolor{blk_color_1}{rgb}{1.000,0.000,0.365}
\definecolor{blk_color_2}{rgb}{1.000,0.000,0.031}
\definecolor{blk_color_3}{rgb}{0.000,1.000,1.000}
\definecolor{blk_color_4}{rgb}{1.000,0.569,0.000}
\definecolor{blk_color_5}{rgb}{0.667,1.000,0.000}
\definecolor{blk_color_6}{rgb}{0.000,0.031,1.000}
\definecolor{blk_color_7}{rgb}{1.000,0.667,0.000}
\definecolor{blk_color_8}{rgb}{0.498,1.000,0.000}
\definecolor{blk_color_9}{rgb}{0.000,0.933,1.000}
\definecolor{blk_color_10}{rgb}{0.000,1.000,0.667}
\definecolor{blk_color_11}{rgb}{0.933,1.000,0.000}
\definecolor{blk_color_12}{rgb}{0.000,0.933,1.000}
\definecolor{blk_color_13}{rgb}{0.769,1.000,0.000}
\definecolor{blk_color_14}{rgb}{0.000,0.498,1.000}
\definecolor{blk_color_15}{rgb}{1.000,0.169,0.000}
\definecolor{blk_color_16}{rgb}{0.000,1.000,0.733}
\definecolor{blk_color_17}{rgb}{0.769,1.000,0.000}
\definecolor{blk_color_18}{rgb}{0.000,1.000,0.400}
\definecolor{blk_color_19}{rgb}{1.000,0.000,0.667}

\begin{figure}[t]
	\centering
	\resizebox{\linewidth}{!}{
		\begin{tabular}{@{}ccccccc@{}}
			\cellcolor{blk_color_0} building+pole &
			\cellcolor{blk_color_1} road+sidewalk &
			\cellcolor{blk_color_2} road &
			\cellcolor{blk_color_3} sidewalk+building &
			\cellcolor{blk_color_4} building+traffic sign &
			\cellcolor{blk_color_5} building+car &
			\cellcolor{blk_color_6} road+car \\
			\cellcolor{blk_color_7} building &
			\cellcolor{blk_color_8} building+vegetation &
			\cellcolor{blk_color_9} road+pole &
			\cellcolor{blk_color_10} building+sky &
			\cellcolor{blk_color_11} pole+car &
			\cellcolor{blk_color_12} building+person &
			\cellcolor{blk_color_13} pole+vegetation
		\end{tabular}
	}
	\centering
	\includegraphics[width=\linewidth]{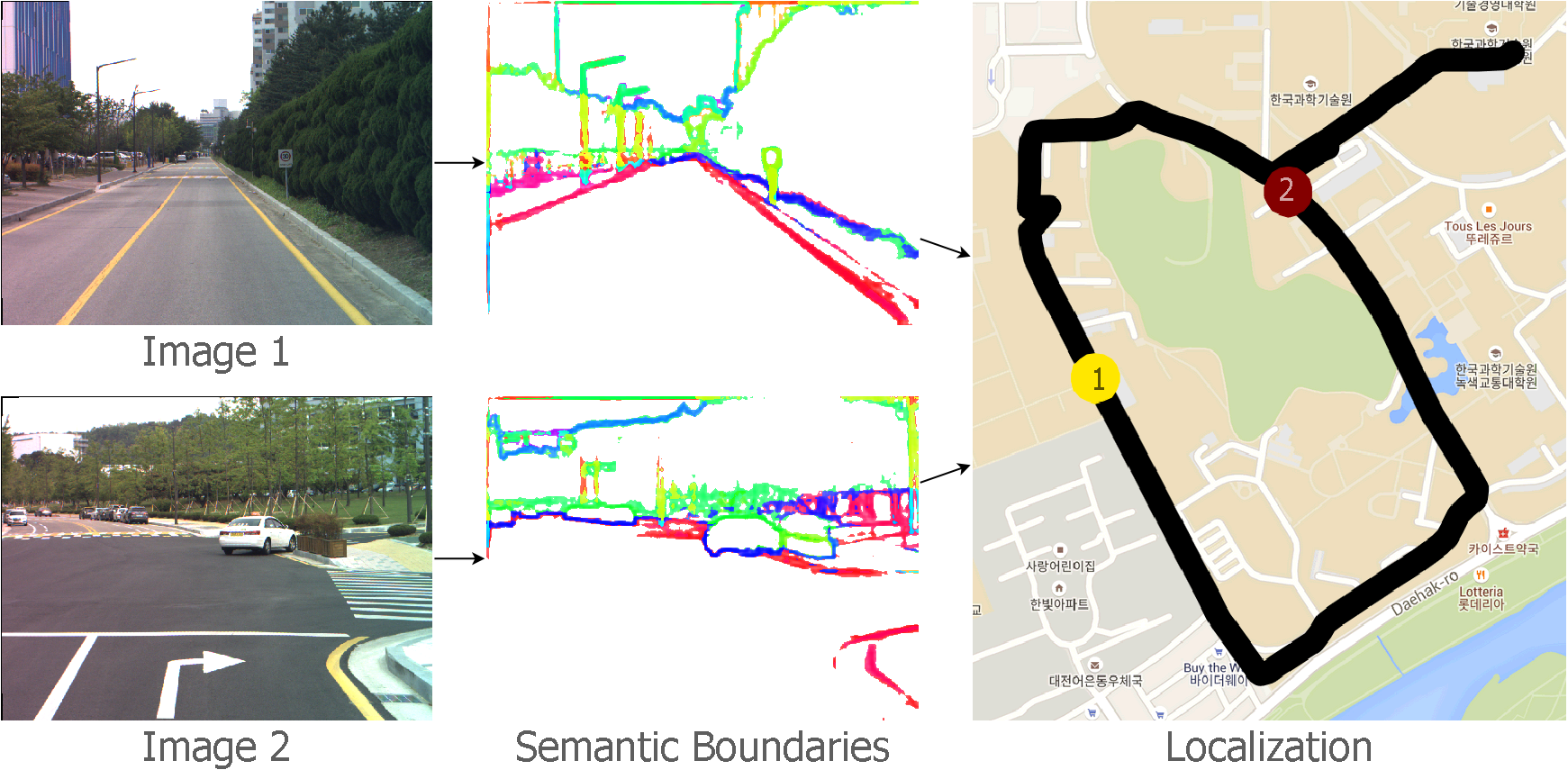}
	\caption{
Illustration of VLASE. Given images (left) from a vehicle, we extract semantic edge features (middle). Different colors indicate different combinations of object classes. The extracted semantic features are compared to the features from geo-tagged images in a database to estimate the location. In this example, the red and yellow circles on the map (right) indicate the locations of the two given images. (The images are from the KAIST WEST sequences captured at 9AM \cite{Choi2015}.).
}
\label{fig:intro_figure}
\end{figure}

Fig~\ref{fig:intro_figure} illustrates the basic idea of this paper. Given an image from a vehicle, we first detect semantic boundaries, the pixels between different object classes. In this paper, we use the recently introduced CASENet~\cite{yu2017casenet} architecture to extract semantic boundaries. The CASENet architecture not only produces state-of-the-art semantic performance on standard datasets such as SBD \cite{hariharan2011semantic} and Cityscapes \cite{cordts2016cityscapes} but also provides a multi-label framework where the edge pixels are associated with more than one object classes. For example, a pixel lying on the edge between sky and buildings will be associated with both sky and building class labels. This allows our method to unify multiple semantic classes as localization features. The middle column of Fig.~\ref{fig:intro_figure} shows the semantic edge features. By matching the semantic edge features between a query image and geo-tagged images in a database, which is achieved using VLAD in this paper, we can estimate the location of the query image, as illustrated on the Google map in the right of Fig.~\ref{fig:intro_figure}.

Besides the matching between semantic edge features, we also observed that in the context of on-road vehicles, appending 2D spatial location information with the extracted features (SIFT or CASENet) boosts the localization performance by a large margin. In this paper, we heavily rely on the prior that the images are captured from a vehicle-mounted camera, and exploit edge features that are typical in urban scenes. In addition, we sample only a very limited set of poses for on-road vehicles. The motion is near-planar and the orientation is usually aligned with the direction of the road. It is common for many recent methods to make this assumption since the primary application is the accurate vehicle localization in urban canyons, where GPS suffers from multi-path effects.

We briefly summarize our contributions as follows:

\begin{itemize}

\item We propose VLASE, a simple method that uses semantic edge features for the task of vehicle localization. The idea of simple semantic cues for localization is not completely new, as individual features such as horizon, road maps, and skylines~\cite{Bansal2014,meguro07,ramalingam10,Sourer2015} have been shown to be beneficial. In contrast to these methods, our method is a unified framework that allows the incorporation of multiple semantic classes.

\item We show that it is beneficial to augment semantic features by 2D spatial coordinates. This is counter-intuitive to prior methods that utilize invariant features in a bag-of-words paradigm. In particular, we show that even standard SIFT-VLAD can be significantly improved by embedding additional keypoint locations.

\item We show compelling experimental results on two different datasets, including the public KAIST~\cite{Choi2015} and a route collected by us in Salt Lake City. 
We outperform competing localization methods such as standard SIFT-VLAD \cite{Jegou12PAMI}, pre-trained NetVLAD~\cite{arandjelovic2016netvlad}, and the coarse localization in~\cite{toft2017long}, even with smaller descriptor dimensions. Our results are comparable and probably slightly better than the improved SIFT-VLAD that incorporates keypoint locations in the features.

\end{itemize}

\section{Related Work}

The vision~\cite{sattler2017benchmarking} and robotics~\cite{lowry2016visual} communities have witnessed the rise of accurate and efficient image-based localization techniques that can be complementary to GPS, which are prone to error due to multi-path effects. The techniques can be classified into regression-based methods and retrieval-based ones. Regression-based methods~\cite{kendall2015posenet,weyand2016planet,brachmann2017dsac} directly obtain the location coordinates from a given image using techniques such as CNNs. Retrieval-based methods match a given query image to thousands of geo-tagged images in a database, and predict the location estimates for the query image based on the nearest or k-nearest neighbors in the database. Regression-based methods provide the best advantage in both memory and speed. For example, methods like PoseNet~\cite{kendall2015posenet} does not require huge database with millions of images and the location estimation can be done in super-real time (e.g. 200 Hertz). On the contrary, retrieval-based ones are usually slower and have a large memory requirement for storing images or its descriptors for the entire city of globe. However, the retrieval-based methods typically provide higher accuracy and robustness~\cite{sattler2017benchmarking}.

\subsection{Features}

In this paper, we will focus on the retrieval-based approach, which essentially find the distance between a pair of images using extracted localization features. Based on human understandability, we broadly classify the localization features into the following two categories: 


\noindent
{\bf Simple Features:}
We refer to simple features as the ones that are human-understandable: line-segments, horizon, road maps, and skylines. Skylines or horizon separating sky from buildings or mountains can be used for localiation~\cite{Bansal2014,meguro07,ramalingam10,Sourer2015}. Several existing methods use 3D models and/or omni-directional cameras for geolocalization~\cite{koch07,stein95,meguro07,tardif08,Zeisl2015,Sattler2012,Yunpeng2012,Torii2013,Majdik:2015}. Line segments have been shown to be very useful for localization. The localization can be achieved by registering an image with a 3D model or a geo-tagged image. By directly aligning the lines from query images to the ones in a line-based 3D model we can achieve localization~\cite{koch07,Ramalingam2011,Micusik2015}. Semantic segmentation of buildings has been used for registering images to 2.5D models~\cite{Armagan2017}.

We can also use other human-understandable simple feature such as roadmaps or weather patterns to obtain localization. Visual odometry can provide the trajectory of a vehicle in motion, and by comparing this with the roadmaps, we can compute the location of the vehicle~\cite{Badino2012,Brubaker2013}. It is intriguing to see that even weather patterns can act as signatures for localizing an image~\cite{jacobs07}. 

\noindent
{\bf Complex Features:}
The complex ones are visual patterns extracted through hand-crafted feature descriptors or automatically extracted ones using CNNs. These class of features are referred to as complex ones since they are not human-understandable, i.e, not easily perceptible to human eye. One of the earlier methods used SIFT or SURF descriptors to match a query image with a database of images~\cite{robertson04,zhang06,Cummins2011}. It is possible to achieve localization in a global scale using GPS-tagged images from the web and matching the query image using a wide variety of image features such as color and texton histograms, gist descriptor, geometric context, and even timestamps~\cite{hays08,kalogerakis09}. 


The use of neural networks for localization is an old idea. RATSLAM~\cite{Milford2004} is a classical SLAM algorithm that uses a neural network with local view cells to denote locations and pose cells to denote heading directions. The algorithm produces ``very coarse'' trajectory in comparison to existing SLAM techniques that employ filtering methods or bundle-adjustment machinery. Kendall~\etal~\cite{kendall2015posenet} presented PoseNet, a 23 layer deep convolutional neural network based on GoogleNet~\cite{szegedy2015going}, to compute the pose in a large-region at 200~Hz. 
CNN can be also applied to learn the distance metric to match two images. As one can achieve localization by matching an image taken at the ground level to reference database of geo-tagged bird's eye, aerial, or even satellite images~\cite{tian2017cross,vo2016localizing,lin2015learning,Pillai2017}, such cross-matching is typically done using siamese networks~\cite{bromley1994signature}. 
Recently, it was shown that LSTMs can be used to achieve accurate localization in challenging lighting conditions~\cite{WalchHLSHC16}. A survey of different state-of-the-art localization techniques is given in ~\cite{lowry2016visual}, and there has been releases of many newer datasets~\cite{sun2017dataset,sattler2017benchmarking}. The idea of dominant set clustering is powerful for localization tasks~\cite{zemene2017large}. 
Many existing methods formulate localization problem in a similar manner to per-exemplar SVMs in object recognition. To handle the limitation of having very few positive training examples, a new approach to calibrate all the per-location SVM classifiers using only the negative examples is proposed~\cite{gronat2013learning}.

In this paper, we combine the above two categories by localizing from human-interpretable semantic edge features learnt from a state-of-the-art CNN~\cite{yu2017casenet}. Note that very recently semantic segmentation is also used with either a sparse 3D model~\cite{toft2017long} or depth images~\cite{schonberger2017semantic} for long-term 3D localization. We show by experiments that VLASE improves the semantic-histogram-based coarse localization in~\cite{toft2017long}.

\subsection{Vocabulary tree}

In the retrieval based methods, we match a query image to millions of images in a database. The computation efficiency is largely addressed by bag-of-words (BOW) representation that aggregates local descriptors into a global descriptor, and enables fast large-scale image search~\cite{Nister2006,Schindler2007,Lee2014}. Recently, extensions of BOW including the Fisher vector and Vector of Locally Aggregated Descriptors (VLAD) showed state-of-the-art performance~\cite{Jegou12PAMI}. Experimental results demonstrate that VLAD significantly outperforms BOW for the same size. It is cheaper to compute and its dimensionality can be reduced to a few hundreds of components by PCA without noticeably impacting its accuracy.

The logical extension to VLAD it NetVLAD, where Arandjelovi{\'c} ~\etal propose to mimic VLAD in a CNN framework and design a trainable generalized VLAD layer, NetVLAD, for the place recognition task~\cite{arandjelovic2016netvlad}. This layer can be used in any CNN architecture and allows training via backward propagation. NetVLAD was shown to outperform non-learnt image representations and off-the-shelf CNN descriptors on two challenging place recognition benchmarks, and improves over current state of-the-art compact image representations on standard image retrieval benchmarks.


%

\section{Semantic Edges for Localization}

This section explains our main algorithm of using semantic edge features for localization. The main idea is very simple. Similar to the use of SIFT features in a VLAD framework, we use CASENet multi-label semantic edge class probabilities as compact, low-dimensional, and interpretable features. Similar to standard BOW, VLAD also constructs a codebook from a databse of feature descriptors (SIFT or CASENet) by performing a simple K-means clustering algorithm on those descriptors. Here we denote $M$ clusters as ${\cal C} = \{c_1,\dots,c_M\}$. Given a query image, each of its feature descriptors $x_i$ is associated to the nearest cluster $c_j$ in the codebook. The main idea in VLAD is to accumulate the difference vector $x_i - c_j$ for every $x_i$ that is associated with $c_j$. VLAD is considered to be superior to traditional BOW methods mainly because this residual statistic provides more information and enables better discrimination.

\definecolor{blk_color_0}{rgb}{1.000,0.400,0.000}
\definecolor{blk_color_1}{rgb}{1.000,0.000,0.365}
\definecolor{blk_color_2}{rgb}{1.000,0.000,0.031}
\definecolor{blk_color_3}{rgb}{0.000,1.000,1.000}
\definecolor{blk_color_4}{rgb}{1.000,0.569,0.000}
\definecolor{blk_color_5}{rgb}{0.667,1.000,0.000}
\definecolor{blk_color_6}{rgb}{0.000,0.031,1.000}
\definecolor{blk_color_7}{rgb}{1.000,0.667,0.000}
\definecolor{blk_color_8}{rgb}{0.498,1.000,0.000}
\definecolor{blk_color_9}{rgb}{0.000,0.933,1.000}
\definecolor{blk_color_10}{rgb}{0.000,1.000,0.667}
\definecolor{blk_color_11}{rgb}{0.933,1.000,0.000}
\definecolor{blk_color_12}{rgb}{0.000,0.933,1.000}
\definecolor{blk_color_13}{rgb}{0.769,1.000,0.000}
\definecolor{blk_color_14}{rgb}{0.000,0.498,1.000}
\definecolor{blk_color_15}{rgb}{1.000,0.169,0.000}
\definecolor{blk_color_16}{rgb}{0.000,1.000,0.733}
\definecolor{blk_color_17}{rgb}{0.769,1.000,0.000}
\definecolor{blk_color_18}{rgb}{0.000,1.000,0.400}
\definecolor{blk_color_19}{rgb}{1.000,0.000,0.667}

\begin{figure}
	\centering
	\resizebox{0.46\textwidth}{!}{
		\begin{tabular}{@{}ccccccc@{}}
			\cellcolor{blk_color_0} building+pole &
			\cellcolor{blk_color_1} road+sidewalk &
			\cellcolor{blk_color_2} road &
			\cellcolor{blk_color_3} sidewalk+building &
			\cellcolor{blk_color_4} building+traffic sign &
			\cellcolor{blk_color_5} building+car &
			\cellcolor{blk_color_6} road+car \\
			\cellcolor{blk_color_7} building &
			\cellcolor{blk_color_8} building+vegetation &
			\cellcolor{blk_color_9} road+pole &
			\cellcolor{blk_color_10} building+sky &
			\cellcolor{blk_color_11} pole+car &
			\cellcolor{blk_color_12} building+person &
			\cellcolor{blk_color_13} pole+vegetation
		\end{tabular}
	}
	\centering
	\includegraphics[width=.47\columnwidth]{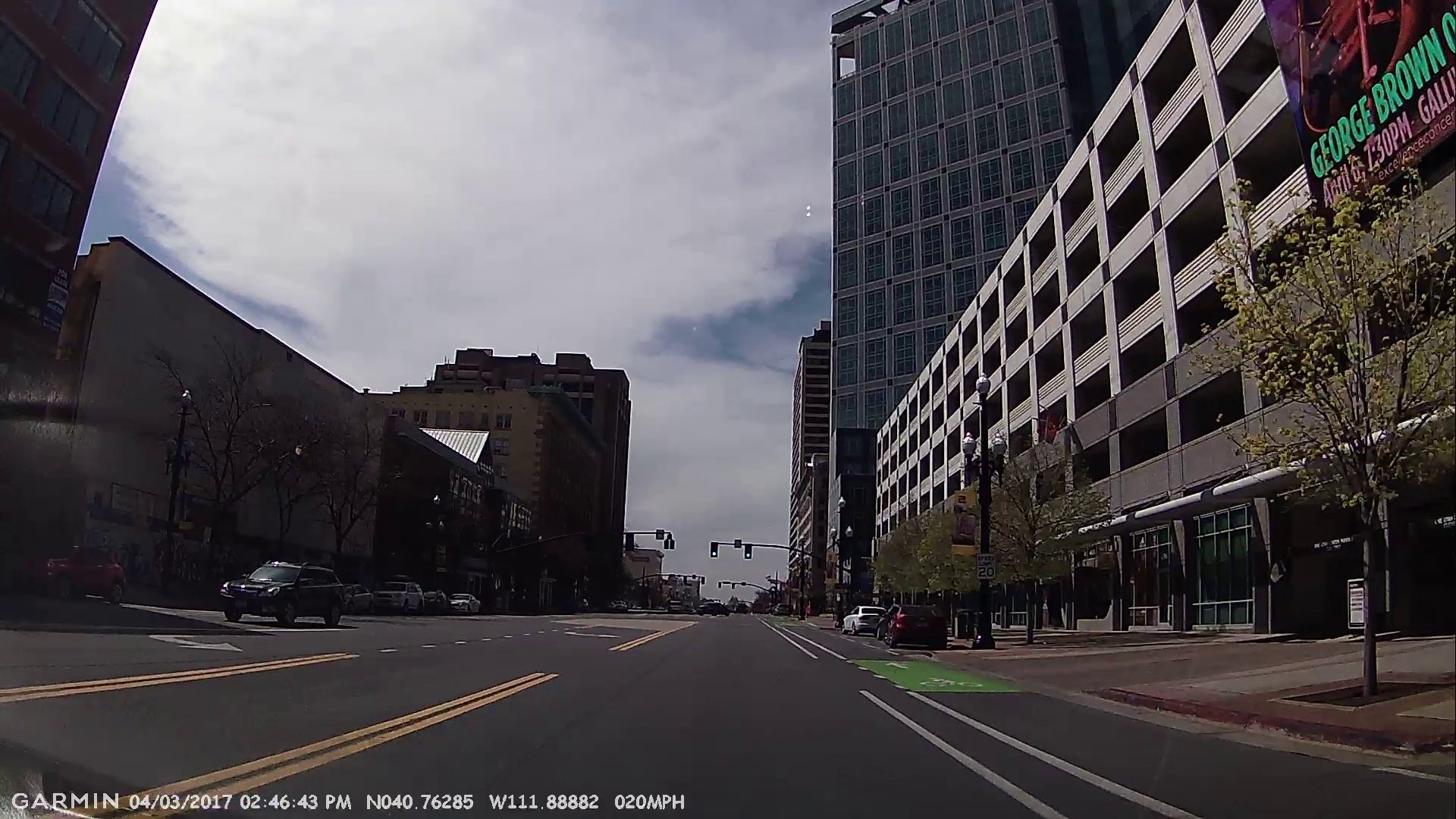}
	\includegraphics[width=.47\columnwidth]{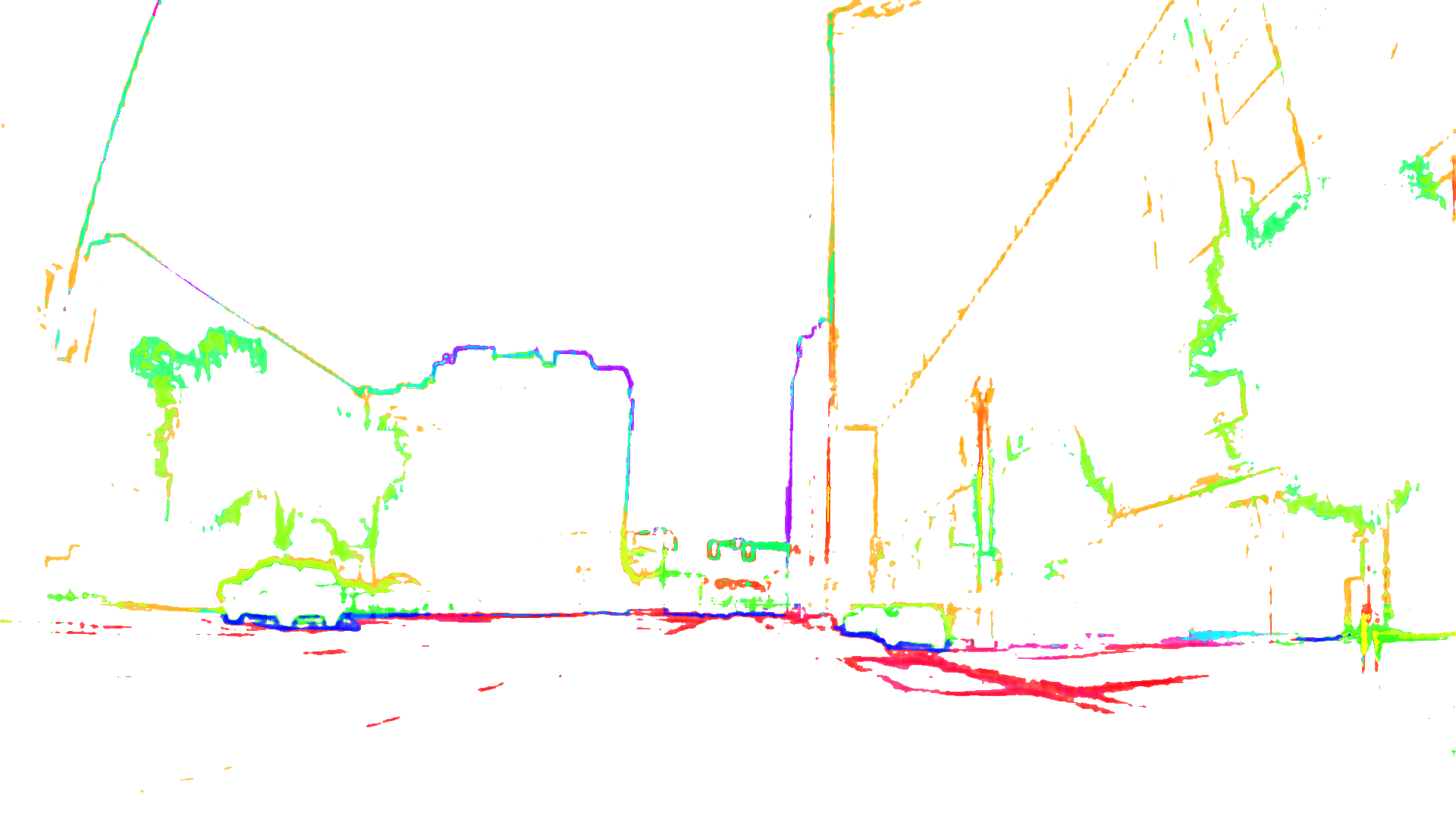}
	\includegraphics[width=0.95\columnwidth]{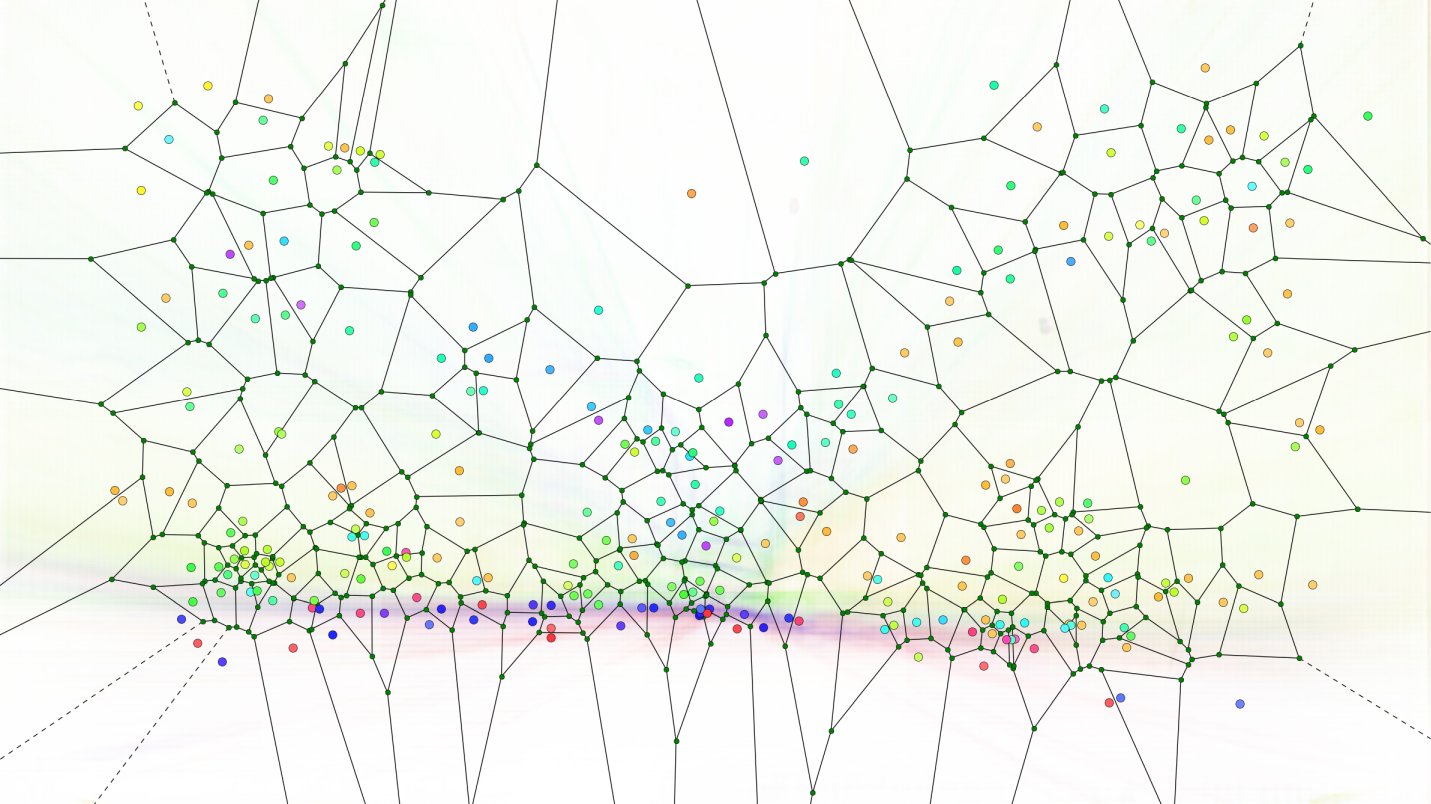}
	\caption{
CASENet edge feature and VLAD. Top: An example input image (left) and its CASENet features (right). Each color corresponds to an object class. Bottom: Visualization of a CASENet-VLAD vocabulary of $M=256$ codewords/cluster centers, shown as color-coded dots. For the dot of each cluster, its x-y positions correspond to $\hat{\mathbf{Y}}_{20}$ and $\hat{\mathbf{Y}}_{21}$, and its color is computed from CASENet features $\mathbf{Y}_k$. The Voronoi graph (black edges with small green nodes) shows the CASENet-VLAD division of the x-y image space. The background of the bottom image is an average of CASENet feature visualization from all images used to train the codebook. As the background shows an averaged semantic on-road driving scene, it can been seen that the colors of the dots in the cluster centers distribute similarly to the colors of this average scene.
}  
\label{fig:casenet-vlad-codebook}
\end{figure}

\begin{figure*}[!t]
	\centering
	\includegraphics[width=0.90\textwidth, trim=0.3in 1.06in 0.6in 0.9in, clip]{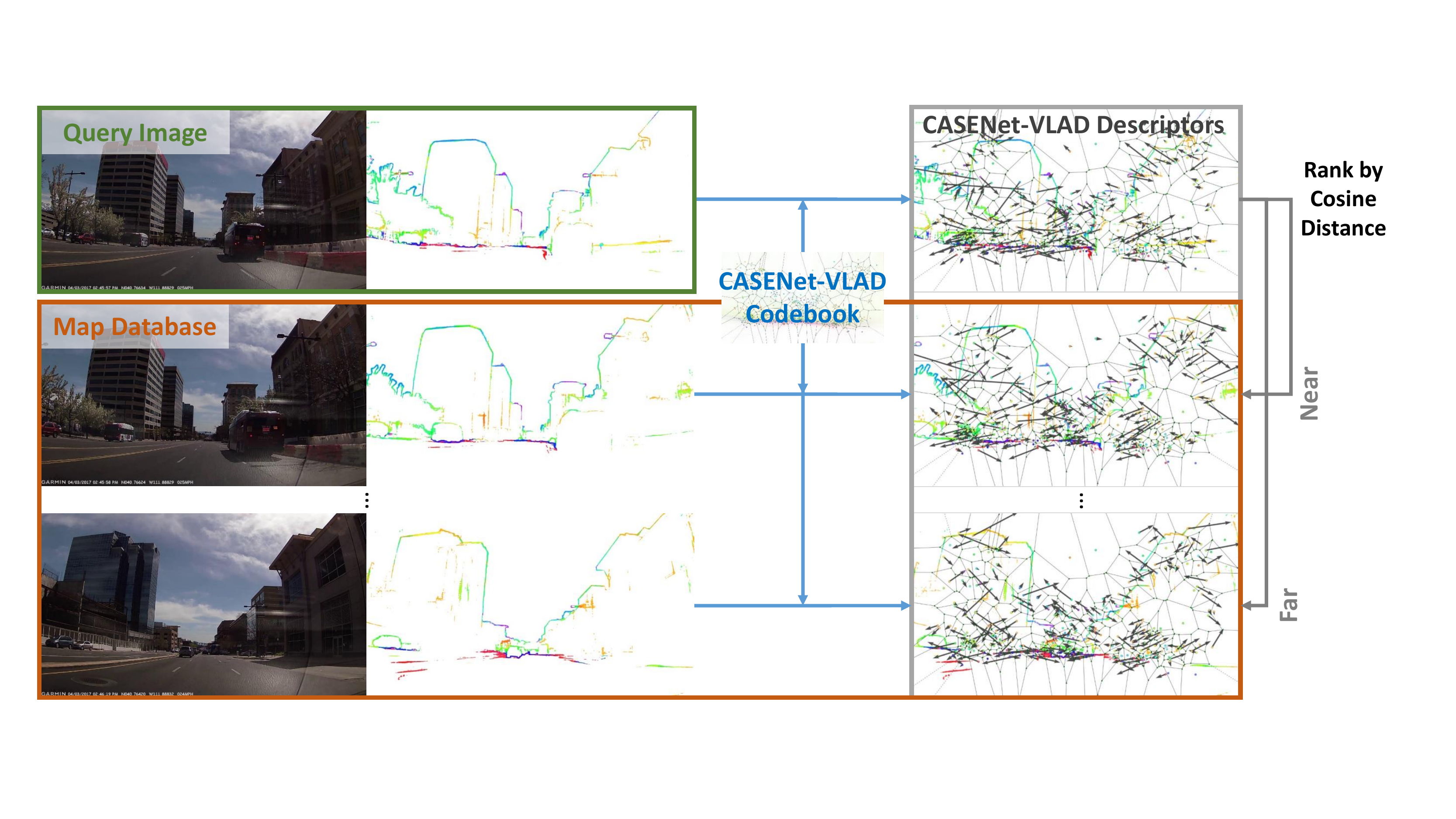}
	\caption{
VLASE pipeline. All mapping images are first processed by CASENet, from which we can build a VLAD codebook using all CASENet features. We then compute each image's CASENet-VLAD descriptor $\mathbf{D}$ (the last two dimensions of each residual vector, i.e., $\mathbf{D}(:,20)$ and $\mathbf{D}(:,21)$, are visualized as 2D vectors origin at the corresponding codeword/cluster center, i.e., $\mathbf{C}_m$). During localization, we similarly compute the currently observed image's CASENet-VLAD descriptor, and query in the database for the top-N closest descriptors in terms of cosine distance. Note that while the geometry shape of the three CASENet edges in column two are visually similar to each other, their corresponding CASENet-VLAD descriptors in the last column are more discriminative, even only visualized by the last two dimensions.
}
\label{fig:pipeline}
\end{figure*}


To detect the semantic edges, we use the recently introduced CASENet architecture, whose code is publicly available~\footnote{\url{http://www.merl.com/research/?research=license-request\&sw=CASENet}}. Given an input image $\mathbf{I}$, we first apply a pretrained CASENet to compute the multi-label semantic edge probabilities $\mathbf{Y(p)=[Y_1(p),\cdots,Y_K(p)]}$ for each pixel $\mathbf{p}\in \mathbf{I}$. Here $K$ is the number of object classes. Then we select all edge pixels $\{ \mathbf{q \in I} | \mathbf{Y}_k(\mathbf{q}) \geq T_e, \exists k \in [1,\cdots,K] \}$, i.e., pixels that have at least one semantic edge label probability exceeding a given threshold $T_e$. Thus, for any image, we can compute a set of $K$-dimensional CASENet edge features (for the Cityscapes dataset, $K = 19$). We further augment this $K$-dimensional feature by appending to its end a 2-dimensional normalized-pixel-position feature $[q_x/W, q_y/H]$, where $W$ and $H$ are the fixed image width and height, and $q_x$ and $q_y$ are the column and row index respectively for a pixel $\mathbf{q}$. We will refer to such a $K + 2$ dimensional feature $\hat{\mathbf{Y}}$ as an augmented CASENet edge feature.

Due to the often much larger number of edge pixels compared to SIFT/SURF features in an image, to build a visual codebook or vocabulary following the VLAD framework, we run a sequential instead of a full KMeans algorithm (MiniBatchKMeans, implemented in the python package scikit-learn~\cite{:/journal/jmlr/2011/pedregosa_scikit-learn}) using all the augmented CASENet edge features on one training image as a mini-batch. This is iterated over the whole training image set for multiple epochs until it converges to $M$ centers $[\mathbf{C}_1,\cdots,\mathbf{C}_M]$, each in the $K + 2$ dimensional space, to form the trained CASENet-VLAD codebook. An example is visualized in~Figure~\ref{fig:casenet-vlad-codebook}.

To perform on-road place recognition, we first need to process a sequence of images serving as the visual map, i.e., the mapping sequence. This can be simply done by extracting all augmented CASENet edge features on each image and compute a corresponding $M \times (K+2)$ CASENet-VLAD descriptor $\mathbf{D}$ using the trained codebook, with power-normalization followed by L2-normalization. The CASENet-VLAD descriptors for the mapping sequence are then stored in a database. During place recognition, we repeat this process for the current query image to get its CASENet-VLAD descriptor and search in the mapping database for the top-N most similar descriptors using cosine-distance. This pipeline is further illustrated in Figure~\ref{fig:pipeline}.

\begin{figure}
	\centering
	\begin{tabular}{cc}
		\includegraphics[height=1.5in]{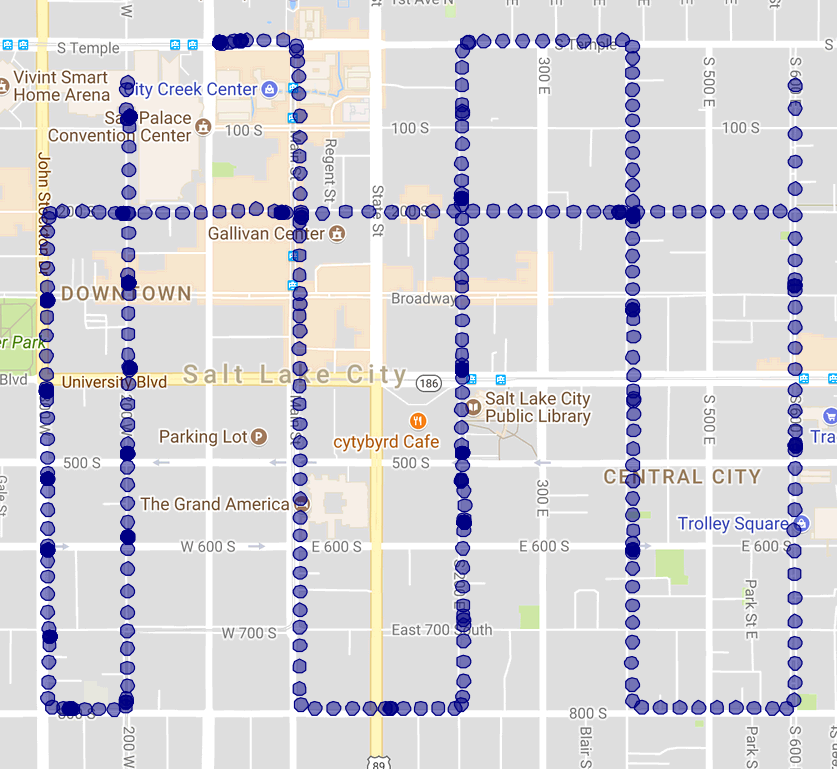} &
		\includegraphics[height=1.5in]{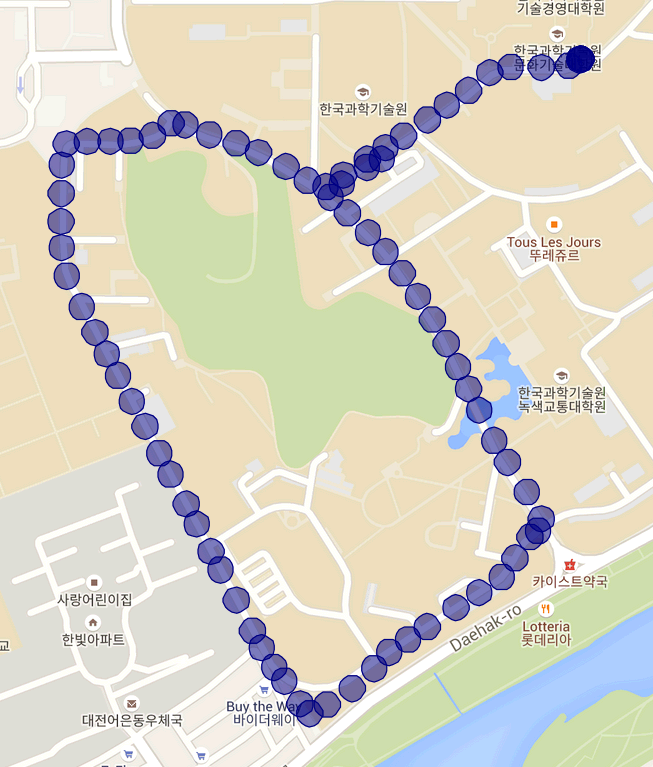} \\
		SLC & KAIST West \\
	\end{tabular}
\caption{
The testing routes of our experiments.
}
\label{fig:routes}
\end{figure}

\section{Datasets}
We have experimented on 2 visual place recognition datasets. The first is called SLC, which was captured in Salt Lake city downtown. The second is called KAIST, which is one of the routes from the KAIST All-Day Visual Place Recognition dataset~\cite{kaistCVPRW2015}. 

\subsection{SLC}


We created our own dataset by capturing two video sequences in the downtown of Salt Lake City. The length of our route is about 15km, which is shown in Figure~\ref{fig:routes}. The two sequences were captured at different times, and thus they have adequate lighting variations for same locations with abundance of objects belonging to the classes in the Cityscapes dataset. We used a Garmin dash-cam to collect videos of the scenes in front of the vehicle. This dash-cam stored the videos at 30 FPS, and the two sequences have 98513 and 89633 frames. We resized the image from the original resolution $1920 \times 1080$ to $640 \times 360$ pixels. A special feature of this dash-cam is that it also encodes the GPS coordinates in latitude and longitude, which provides the ground truth of our video frames.  
Since the frame rate of SLC sequence is 30 fps but only the first frame within every second has a GPS coordinate, we sampled every 30 frames from SLC sequences. We use the longer sequence of SLC (98513) as the database of 3284 images and computing the VLAD codebook, which is denoted as \emph{loop1} hereafter. The other sequence is denoted as \emph{loop2}, which has 2988 sampled frames for querying.


\subsection{KAIST}

    
The KAIST dataset was captured by Choi \etal \cite{kaistCVPRW2015} in the campus of Korea Advanced Institute of Science and Technology (KAIST). They captured 42 km sequences at 15-100Hz using multiple sensor modalities such as fully aligned visible and thermal devices, high resolution stereo visible cameras, and a high accuracy GPS/IMU inertial navigation system. The sequences covered 3 routes in the campus, which are denoted as west, east and north. Each route has 6 sequences recorded at different times of a day, including day (9 AM, 2 PM), night(10 PM, 2 AM), sunset(7 PM), and sunrise(5 AM). As these sequences capture various illumination conditions, this dataset is helpful for benchmarking under lighting variations. 

We used two sequences captured on the west route, as shown in \ref{fig:routes}. The two sequences were captured on 5 AM and 9 AM, which were under sunrise and daylight conditions, respectively. The sequence at 9AM contains more dynamic class objects than that at 5AM. We resized the images from their original size $1280 \times 960$ to $640 \times 480$ pixels. The images were captured at 15 fps while the GPS coordinates were measured at 10 FPS. 
Similar to SLC, we sampled the route captured on 9AM as the database of 3254 images and computing the VLAD codebook, and the route captured on 5AM for querying (2207 images).

\section{Experiments}

\begin{figure*}[t]
	\begin{tabular}{cccc}
		\hspace*{-0.5cm}%
		\includegraphics[width=0.24\linewidth]{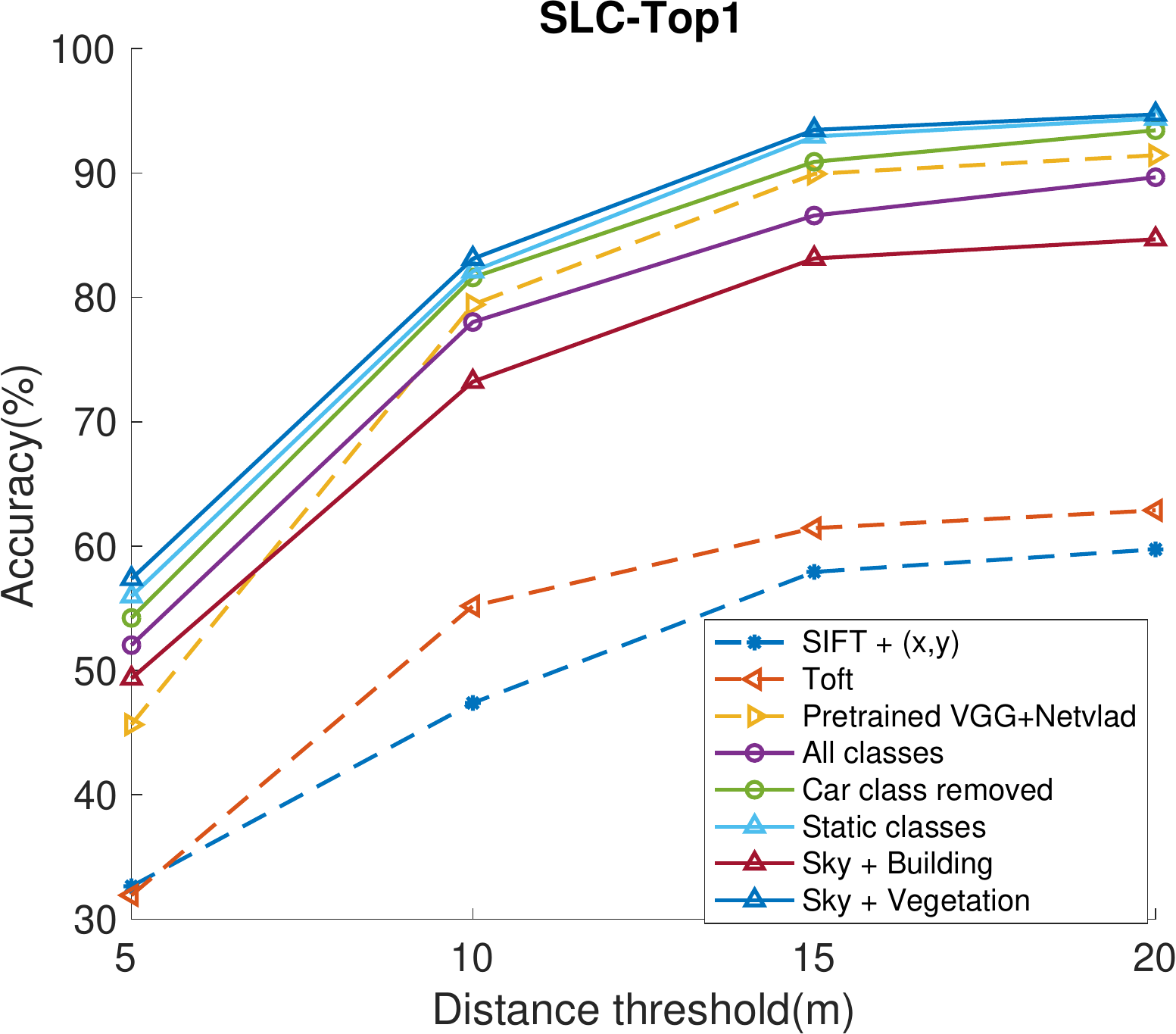} &
		\includegraphics[width=0.24\linewidth]{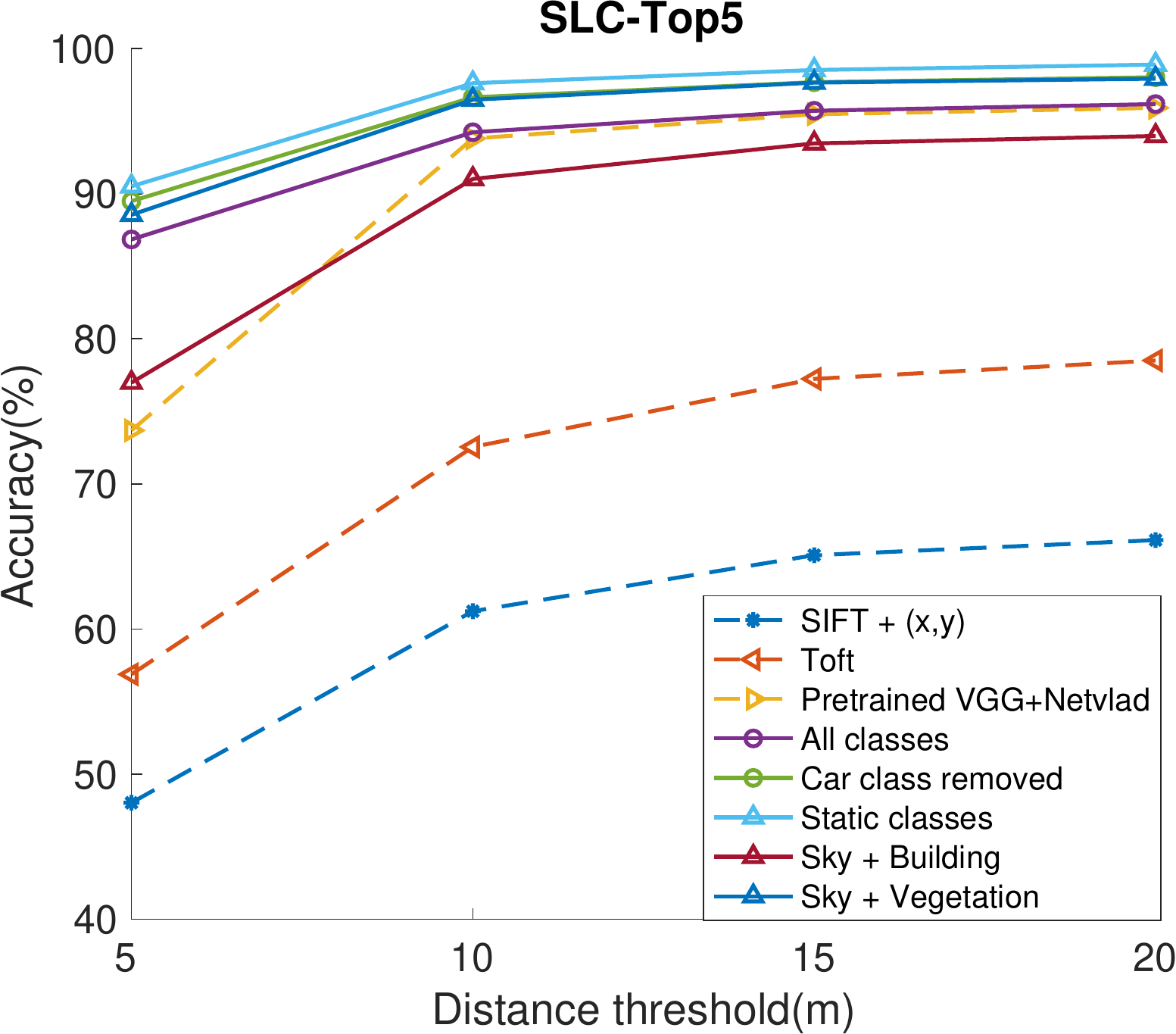} &
		\includegraphics[width=0.24\linewidth]{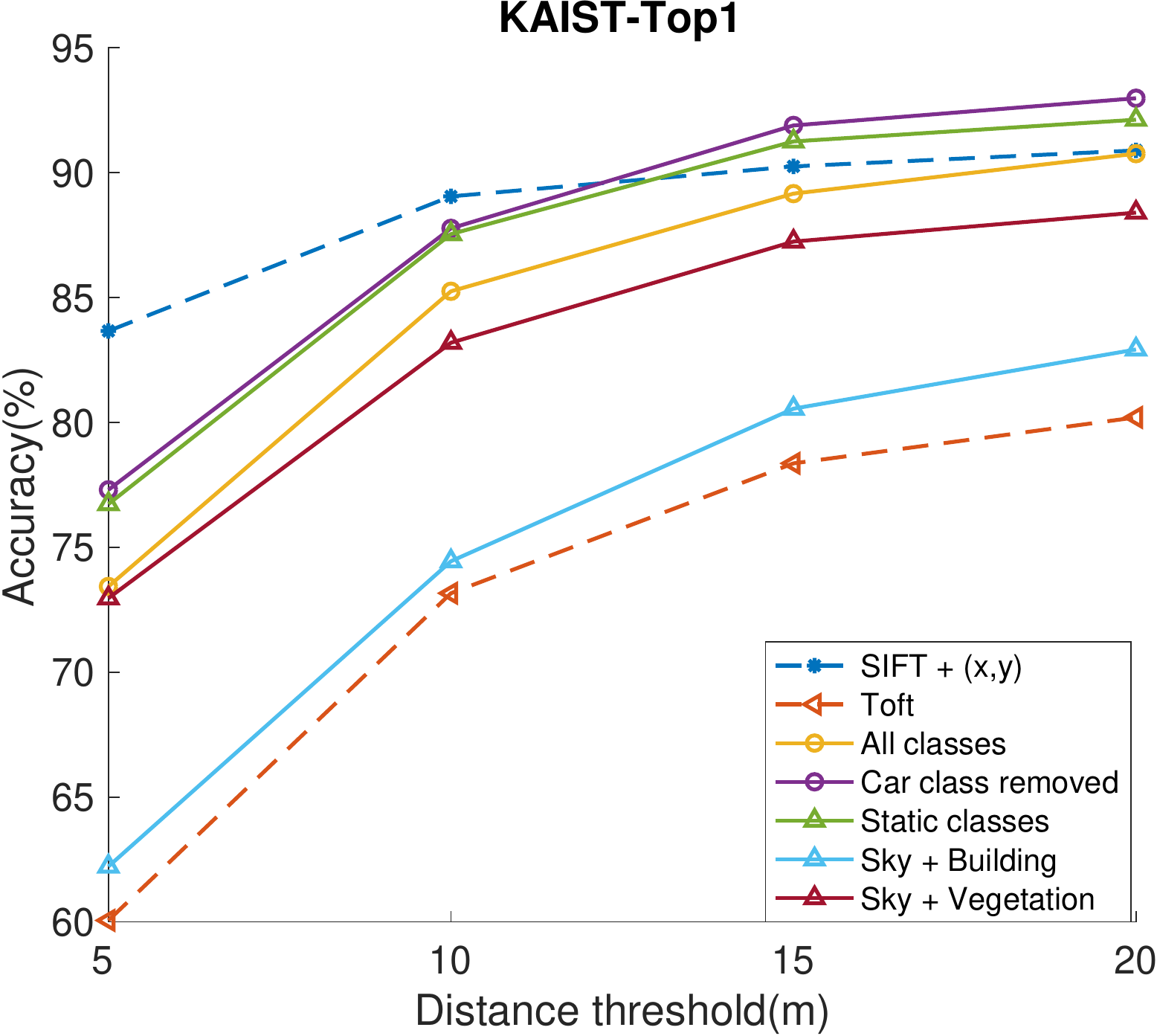} &
		\includegraphics[width=0.24\linewidth]{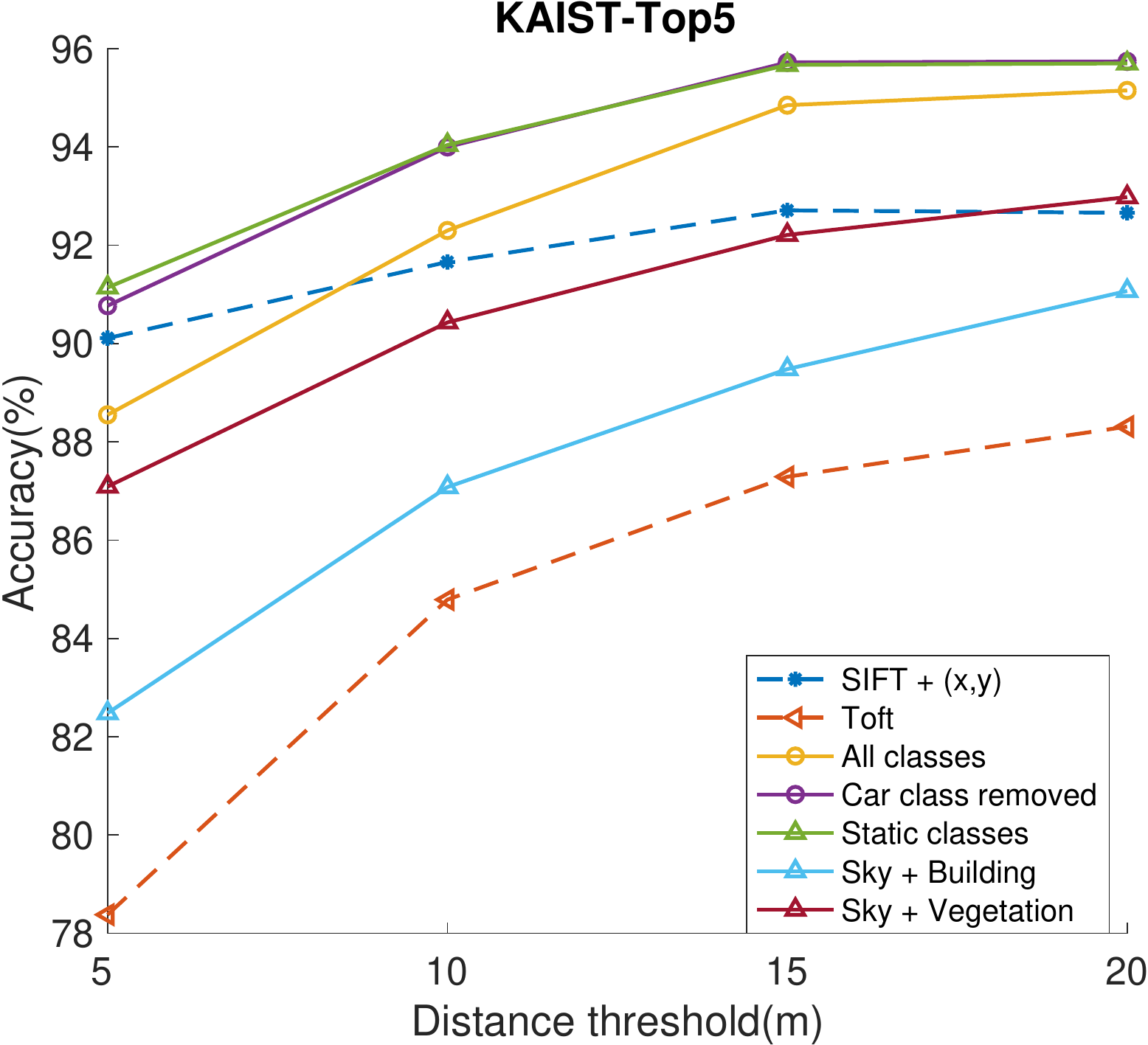} \\
		(a) & (b) & (c) & (d)
	\end{tabular}
	\caption{Localization accuracies. (a) and (b) represent the results for SLC dataset while (c) and (d) represent the results for KAIST dataset. The x-axis represents the distance threshold and the y-axis represents the accuracy. Non-CASENet results are shown using dashed lines. No weighting of features are applied. Note for KAIST, the pretrained VGG-NetVLAD performances are very low (and even with retraining), thus we do not include them here. Note CASENet is not retrained either.\label{fig:accuracy_top1_top5}}
\end{figure*}

\begin{table}[t]
	\caption{
		Ablation study results for the SLC dataset.
	\label{tab:slc}
	}
    \begin{center}
\resizebox{0.75\columnwidth}{!}{
    \begin{tabular}{ |c|c|c|c|c|c|c| } 
    \hline
    & \multicolumn{3}{|c|}{\textbf{Top-1 Accuracy}} & \multicolumn{3}{|c|}{\textbf{Top-5 Accuracy}}\\
    \hline
    \textbf{Removed}
    & \textbf{5m} & \textbf{10m}  & \textbf{20m}
    & \textbf{5m} & \textbf{10m} & \textbf{20m} \\
    \hline
       Road & 53 & 79 & 91 & 89 & 96 & 98\\
   Sidewalk & 54 & 80 & 91 & 87 & 94 & 96\\
   Building & 50 & 75 & 86 & 81 & 90 & 92\\
       Wall & 50 & 76 & 87 & 85 & 92 & 94\\
      Fence & 54 & 80 & 90 & 87 & 94 & 96\\
       Pole & 51 & 75 & 88 & 85 & 93 & 95\\
      Light & 51 & 75 & 87 & 84 & 92 & 95\\
       Sign & 51 & 76 & 87 & 85 & 93 & 95\\
       Veg  & 50 & 74 & 85 & 83 & 92 & 95\\
    Terrain & 51 & 77 & 88 & 84 & 91 & 94\\
        Sky & 50 & 75 & 85 & 82 & 91 & 93\\
        \hline
     Person & 52 & 79 & 90 & 87 & 94 & 96\\
      Rider & 51 & 78 & 89 & 87 & 94 & 96\\
        Car & 54 & 82 & 93 & 89 & 97 & 98\\
      Truck & 53 & 80 & 91 & 88 & 95 & 97\\
        Bus & 51 & 77 & 88 & 84 & 92 & 95\\
      Train & 54 & 79 & 91 & 87 & 95 & 97\\
 Motorcycle & 51 & 77 & 88 & 85 & 92 & 95\\
    Bicycle & 52 & 77 & 89 & 86 & 94 & 96\\
    \hline
    
    \hline\hline
    \textbf{Combinations} & \textbf{5m} & \textbf{10m}  & \textbf{20m} & \textbf{5m} & \textbf{10m} & \textbf{20m} \\
    \hline
        All & 52 & 78 & 90 & 87 & 94 & 96\\
     Static & 56 & 82 & 94 & 91 & 98 & 99\\
    Bld-Sky & 49 & 73 & 85 & 77 & 91 & 94\\
    Veg-Sky & 57 & 83 & 95 & 89 & 96 & 98\\
  Veg-Bld-Sky & 55 & 80 & 91 & 86 & 94 & 96\\
All w/o (x,y) & 44 & 67 & 76 & 77 & 86 & 89\\
    \hline
    
    \hline\hline
    \textbf{Baselines} & \textbf{5m} & \textbf{10m}  & \textbf{20m} & \textbf{5m} & \textbf{10m} & \textbf{20m} \\
    \hline
  SIFT+(x,y) & 32 & 47 & 60 & 48 & 61 & 66\\
       SIFT & 22 & 36 & 43 & 32 & 45 & 48\\
Toft\cite{toft2017long} & 32 & 55 & 63 & 57 & 73 & 79 \\
    \hline
    \end{tabular}
}
    \end{center}
\end{table}

\begin{table}[t]
	\caption{
		Ablation study results for the KAIST dataset.
	\label{tab:kaist}
	}
    \begin{center}
\resizebox{0.75\columnwidth}{!}{
    \begin{tabular}{ |c|c|c|c|c|c|c| } 
    \hline
    & \multicolumn{3}{|c|}{\textbf{Top-1 Accuracy}} & \multicolumn{3}{|c|}{\textbf{Top-5 Accuracy}}\\
    \hline
    \textbf{Removed} & \textbf{5m} & \textbf{10m}  & \textbf{20m} & \textbf{5m} & \textbf{10m} & \textbf{20m} \\
    \hline
       Road & 72 & 84 & 90 & 88 & 91 & 94\\
   Sidewalk & 71 & 84 & 91 & 88 & 92 & 95\\
   Building & 71 & 84 & 90 & 88 & 91 & 94\\
       Wall & 73 & 85 & 90 & 87 & 91 & 94\\
      Fence & 73 & 86 & 92 & 90 & 93 & 96\\
       Pole & 70 & 84 & 89 & 87 & 91 & 94\\
      Light & 73 & 86 & 91 & 88 & 93 & 95\\
       Sign & 71 & 84 & 90 & 88 & 92 & 95\\
        Veg & 69 & 82 & 87 & 87 & 91 & 93\\
    Terrain & 72 & 84 & 90 & 88 & 91 & 94\\
        Sky & 73 & 85 & 91 & 88 & 93 & 95\\
        \hline
     Person & 74 & 86 & 91 & 89 & 92 & 95\\
      Rider & 72 & 85 & 90 & 88 & 92 & 95\\
        Car & 77 & 88 & 93 & 91 & 94 & 96\\
      Truck & 72 & 86 & 90 & 89 & 93 & 94\\
        Bus & 74 & 86 & 90 & 89 & 92 & 94\\
      Train & 74 & 85 & 91 & 88 & 92 & 95\\
 Motorcycle & 72 & 85 & 90 & 88 & 92 & 95\\
    Bicycle & 73 & 85 & 90 & 88 & 92 & 95\\
    \hline
    
    \hline\hline
    \textbf{Combinations} & \textbf{5m} & \textbf{10m}  & \textbf{20m} & \textbf{5m} & \textbf{10m} & \textbf{20m} \\
    \hline
        All & 73 & 85 & 91 & 89 & 92 & 95\\
     Static & 77 & 88 & 92 & 91 & 94 & 96\\
    Bld-Sky & 62 & 74 & 83 & 82 & 87 & 91\\
    Veg-Sky & 73 & 83 & 88 & 87 & 90 & 93\\
  Veg-Bld-Sky & 73 & 84 & 89 & 87 & 91 & 93\\
All w/o (x,y) & 64 & 78 & 85 & 83 & 88 & 91\\
    \hline
    
    \hline\hline
    \textbf{Baselines} & \textbf{5m} & \textbf{10m}  & \textbf{20m} & \textbf{5m} & \textbf{10m} & \textbf{20m} \\
    \hline
 SIFT+(x,y) & 84 & 89 & 91 & 90 & 92 & 93\\
       SIFT & 81 & 86 & 88 & 88 & 89 & 90\\
Toft\cite{toft2017long} & 60 & 73 & 80 & 78 & 85 & 88\\
    \hline
    \end{tabular}
}
    \end{center}
\end{table}

\subsection{Settings}

\noindent \textbf{CASENet}: We use the CASENet model pre-trained on the Cityscapes dataset~\cite{cordts2016cityscapes}. It contains 19 object classes that are also seen in our testing video sequences.
We used nVidia Titan Xp GPUs to extract CASENet features, which can process
around 1.25 images per second using CASENet original code.
We did not retrain CASENet for our datasets, since getting ground truth semantic edges is a tedious manual task. We observed that the pre-trained model was sufficient to provide qualitatively accurate semantic edge features.

\noindent \textbf{VLAD}: We compared the CASENet-based semantic edge features to SIFT~\cite{Jegou12PAMI}, and used VLAD to aggregate both to descriptors for image retrieval. To decide the number of clusters for VLAD, we find the optimal cluster numbers within 32, 64 and 256 by experiments, with MiniBatchKMeans of at most 10,000 iterations. Our experiments showed that 64 clusters for CASENet features and 32 for SIFT are the most optimal, and thus we applied these cluster numbers for further experiments. Note that although CASENet feature dimension is much smaller than SIFT (19 vs. 128), there are more CASENet features for each image as we get them for each pixel. As a result, CASENet works better with more clusters than SIFT. The VLAD of both were trained on CPUs. With Intel(R) Xeon(R) E5-2640 CPU and 125GB of usable memory, the training for 3000 images took about 30 minutes.

\noindent \textbf{Evaluation criteria}: We measured both top-1 and top-5 retrieval accuracy under different distance thresholds (5, 10, 15, and 20 meters). If any of these top-k retrieved images is within the distance threshold of the query image, we counted it as a success localization.

\subsection{Results and Ablation Studies}

\begin{figure*}
	\centering
	\resizebox{0.75\textwidth}{!}{
		\begin{tabular}{@{}ccccccc@{}}
			\cellcolor{blk_color_0} building+pole &
			\cellcolor{blk_color_1} road+sidewalk &
			\cellcolor{blk_color_2} road &
			\cellcolor{blk_color_3} sidewalk+building &
			\cellcolor{blk_color_4} building+traffic sign &
			\cellcolor{blk_color_5} building+car &
			\cellcolor{blk_color_6} road+car \\
			\cellcolor{blk_color_7} building &
			\cellcolor{blk_color_8} building+vegetation &
			\cellcolor{blk_color_9} road+pole &
			\cellcolor{blk_color_10} building+sky &
			\cellcolor{blk_color_11} pole+car &
			\cellcolor{blk_color_12} building+person &
			\cellcolor{blk_color_13} pole+vegetation
		\end{tabular}
	}
	\centering
	\includegraphics[width=0.6\textwidth]{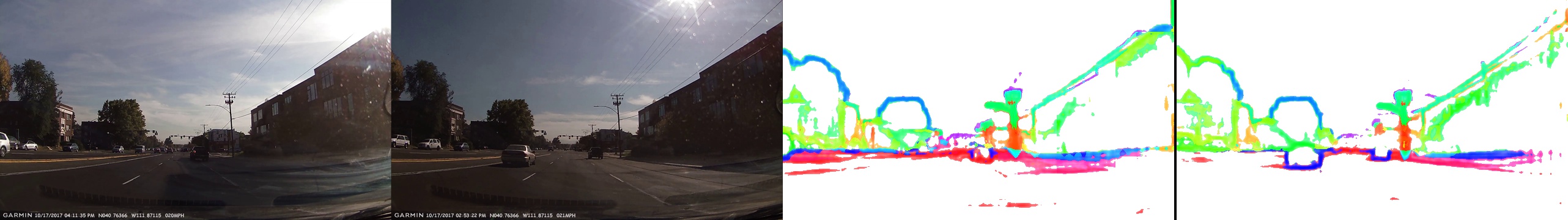}
	\includegraphics[width=0.6\textwidth]{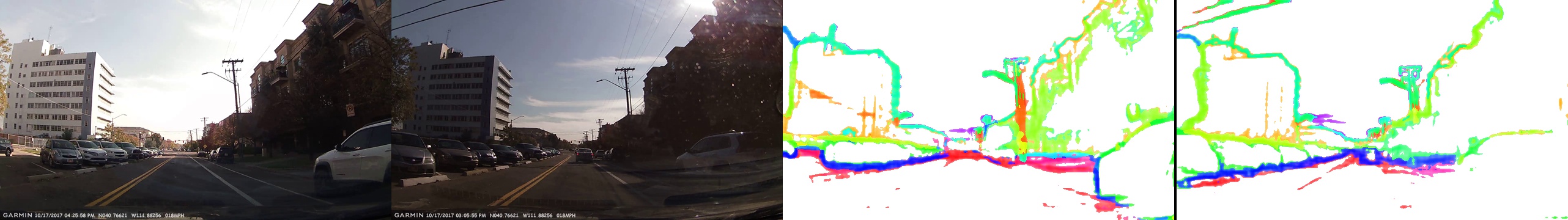}
	\includegraphics[width=0.6\textwidth]{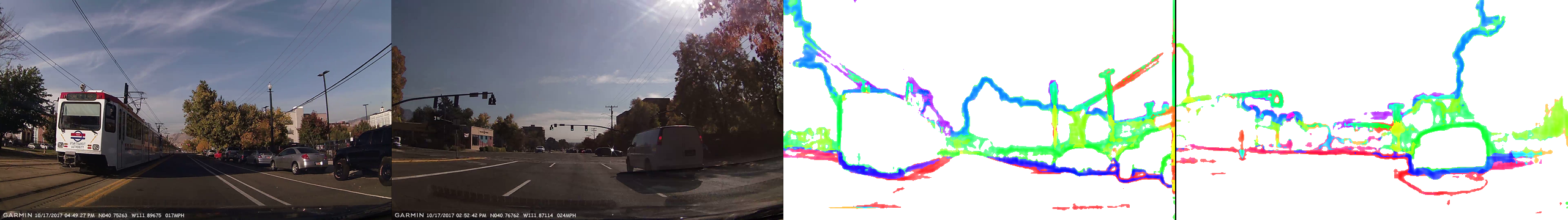}
	\includegraphics[width=0.6\textwidth]{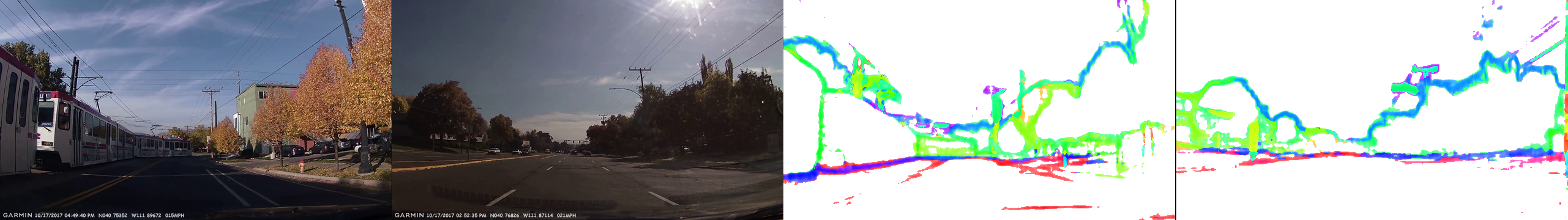}
	\caption{
Successful and failed matches of CASENet+VLAD. The top 2 rows show good matches
. The bottom 2 rows show two of the worst results where the true distance is greater than 2 kms.
In the 3rd row, the presence of dynamic object such as the train might lead to the high error.
}
\label{fig:good_and_bad}
\end{figure*}

Figure~\ref{fig:accuracy_top1_top5} shows our main results compared with several baselines. Fig.~\ref{fig:good_and_bad} presents several best and worst matching examples by our method.
We also performed ablation studies on the importances of 1) object classes and 2) spatial coordinates used for feature augmentation, with results listed in Tables~\ref{tab:slc} and~\ref{tab:kaist}.

\noindent \textbf{Object classes}: We first investigated the importance of different subsets of the 19 Cityscapes classes for localization (all augmented by 2D spatial coordinates) with two goals. The first is to evaluate individual class contributions to the accuracy. The second is to compare our approach with existing methods that also use semantic boundaries but with much fewer classes. For example, one of the popular localization cues is skylines (edges between building and sky, or vegetation and sky)~\cite{Bansal2014,meguro07,ramalingam10,Sourer2015}.

For SLC and in most cases, removing dynamic classes (listed in the second half of the first block of Table~\ref{tab:slc}) yields better accuracy than all classes, e.g., removing cars improves the accuracy by 2\%. Note in some cases, removal of some dynamic classes causes minor drops in accuracy, e.g., removing Motorcycle and bus, which we believe is insignificant, and mainly due to the lack of those classes in our dataset.
As per our expectation, using only static classes (the 11 out of 19 classes) of CASENet performs better than using all classes, for both datasets. Specifically, building, sky and wall are the top 3 individual contributors, as removing them causes highest drop in the accuracy. Also using only vegetation and sky is comparable to using all static classes.

For KAIST, vegetation seems to be the most important individual class. Removing it causes the highest drop in the accuracy. Building and sky classes individually does not seem very significant. Again, using only static CASENet features performs better than any other feature combination. 

\noindent \textbf{Spatial coordinates}:
Besides object classes and their probabilities, we also tried removing the 2D-image-coordinate augmentation from the feature descriptors for both CASENet and SIFT. Surprisingly, this augmentation boosted the performance of both SIFT and CASENet by a large margin: SIFT+(x,y) vs. SIFT, and All vs. All w/o (x,y) in Table~\ref{tab:slc} and~\ref{tab:kaist}. While this result seems counter-intuitive due to the loss of invariance in feature descriptors, the on-road vehicle localization is a more restricted setup and such constraints lead to high-accuracy localization.

\begin{figure*}
	\begin{tabular}{cccc}
		\hspace*{-0.5cm}%
		\includegraphics[width=0.24\linewidth]{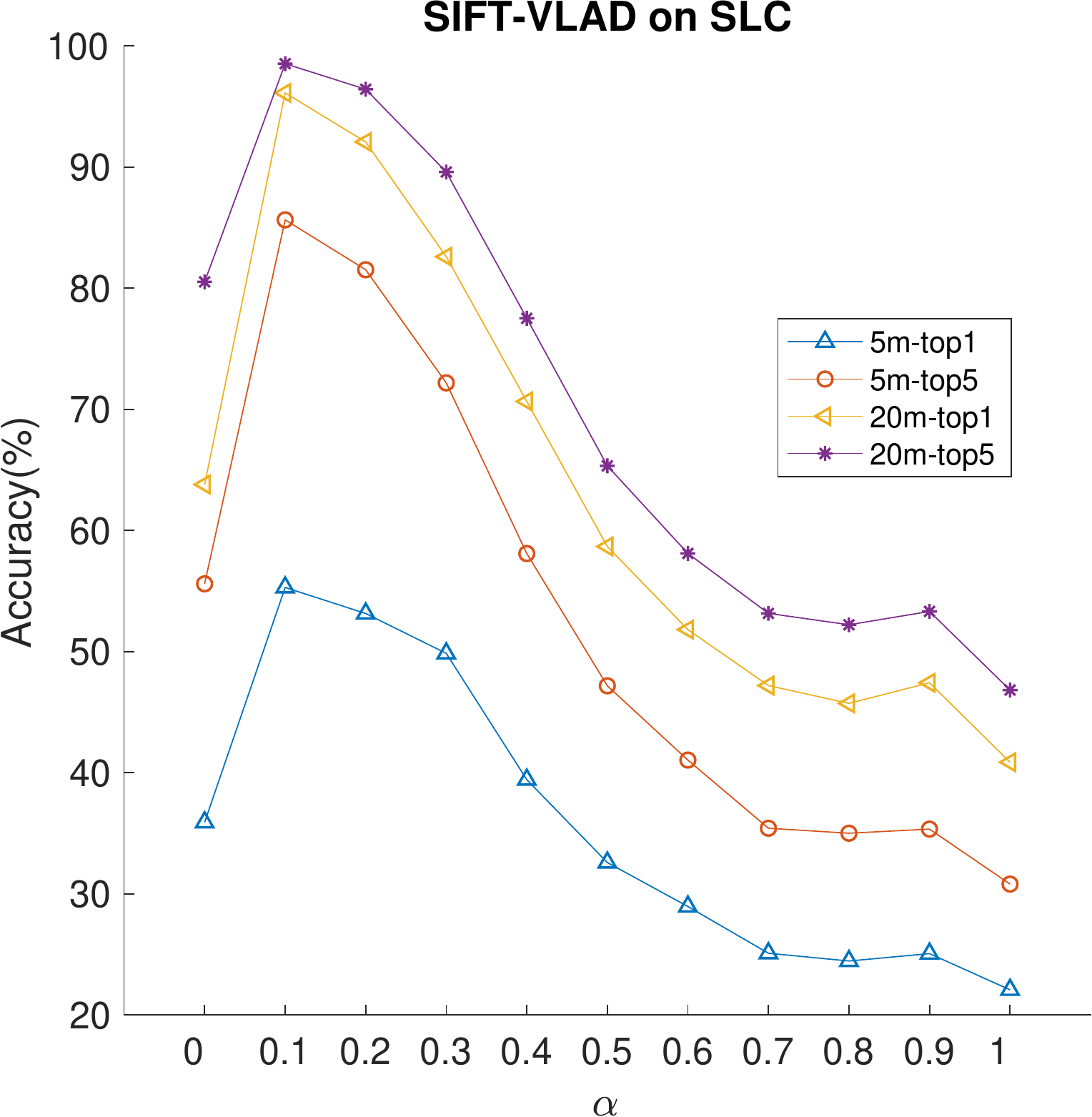} &
		\includegraphics[width=0.24\linewidth]{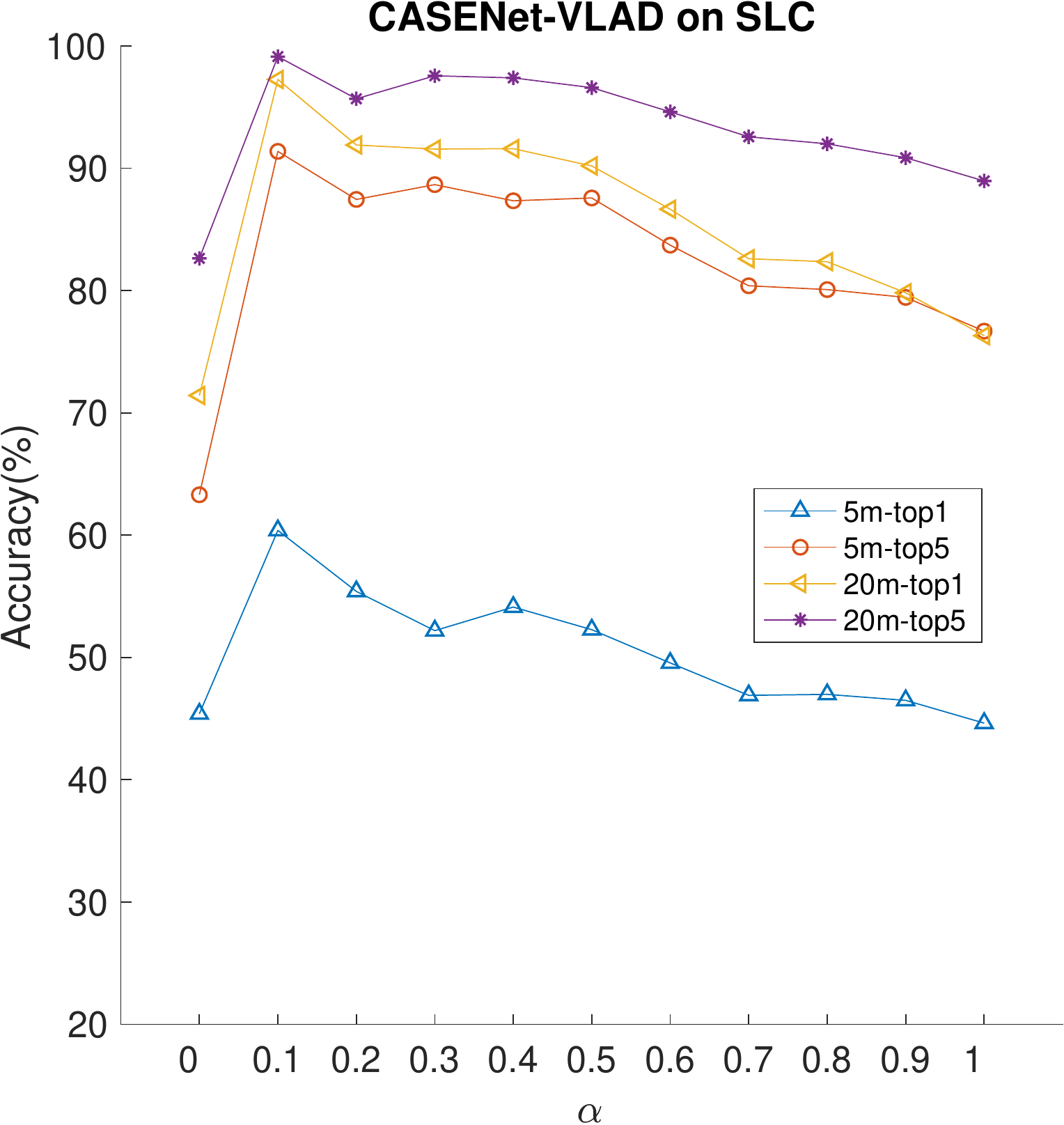} &
		\includegraphics[width=0.24\linewidth]{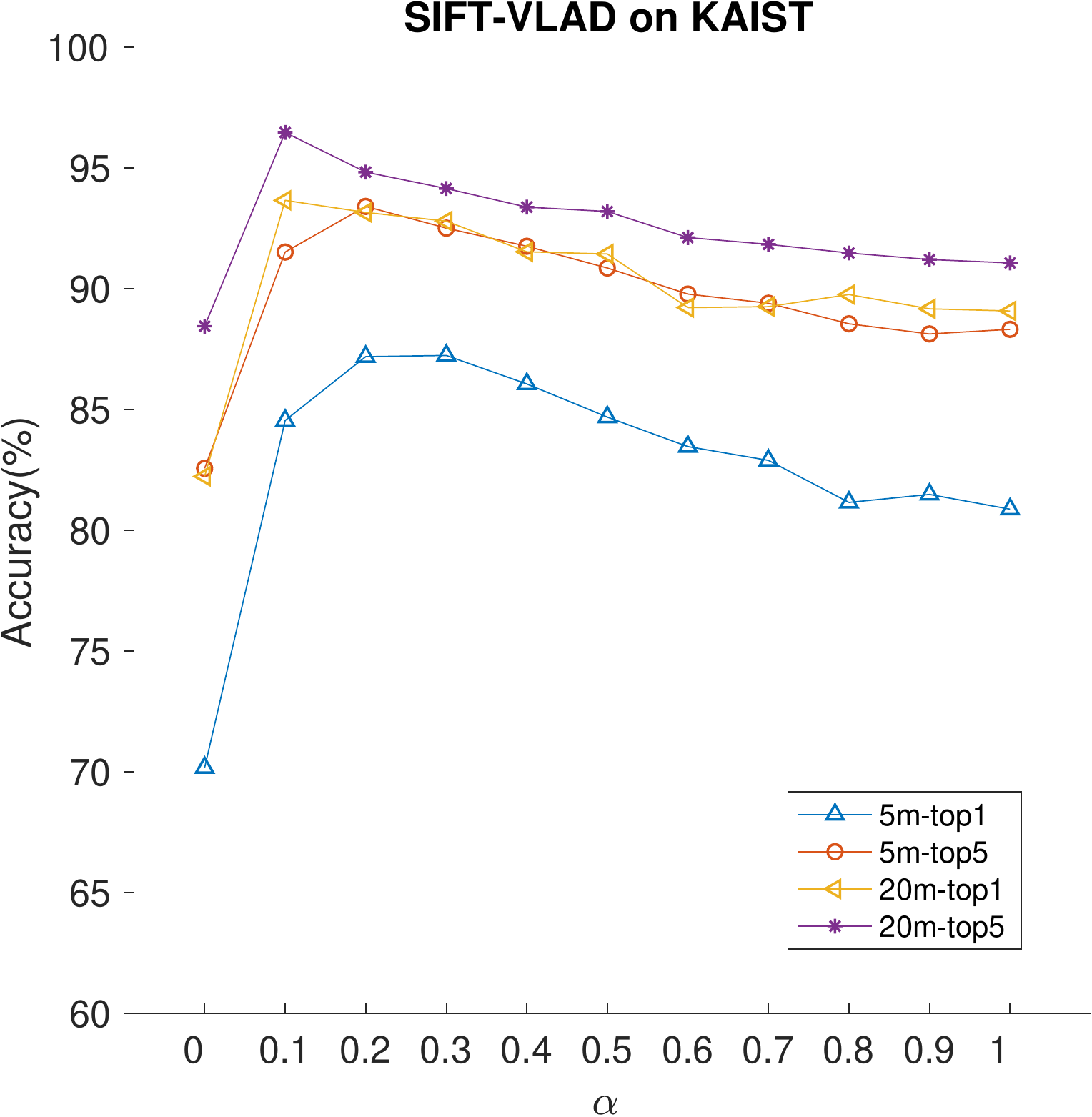} &
		\includegraphics[width=0.24\linewidth]{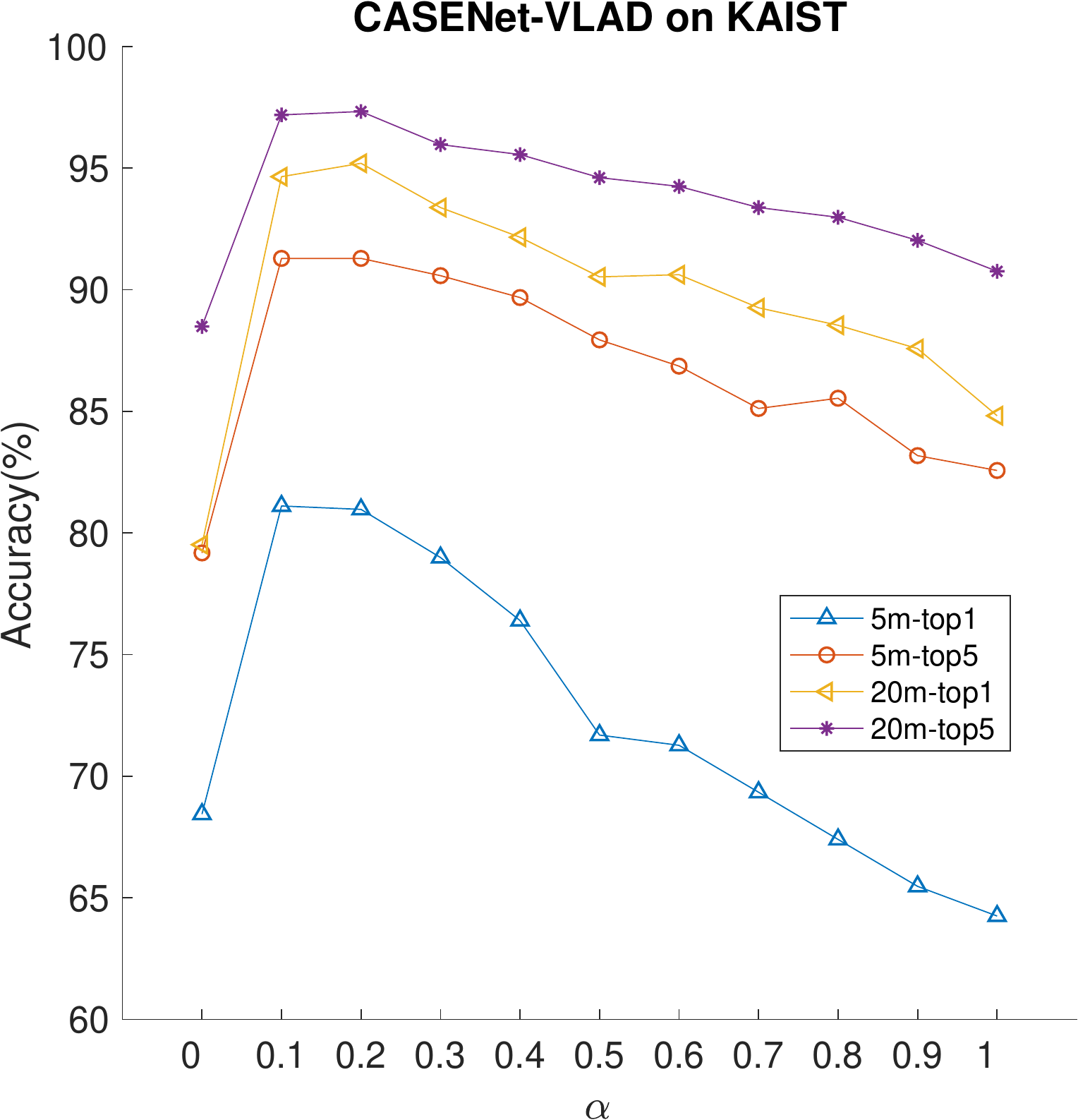}
	\end{tabular}
	\caption{Effect of weighted spatial coordinate augmentation on SLC (left) and KAIST (right). At the optimal $\alpha=0.1$, CASENet is still better than SIFT. \label{fig:alpha-scaling}}
\end{figure*}

\begin{figure*}
	\begin{tabular}{cccc}
		\hspace*{-0.5cm}%
		\includegraphics[width=0.24\linewidth]{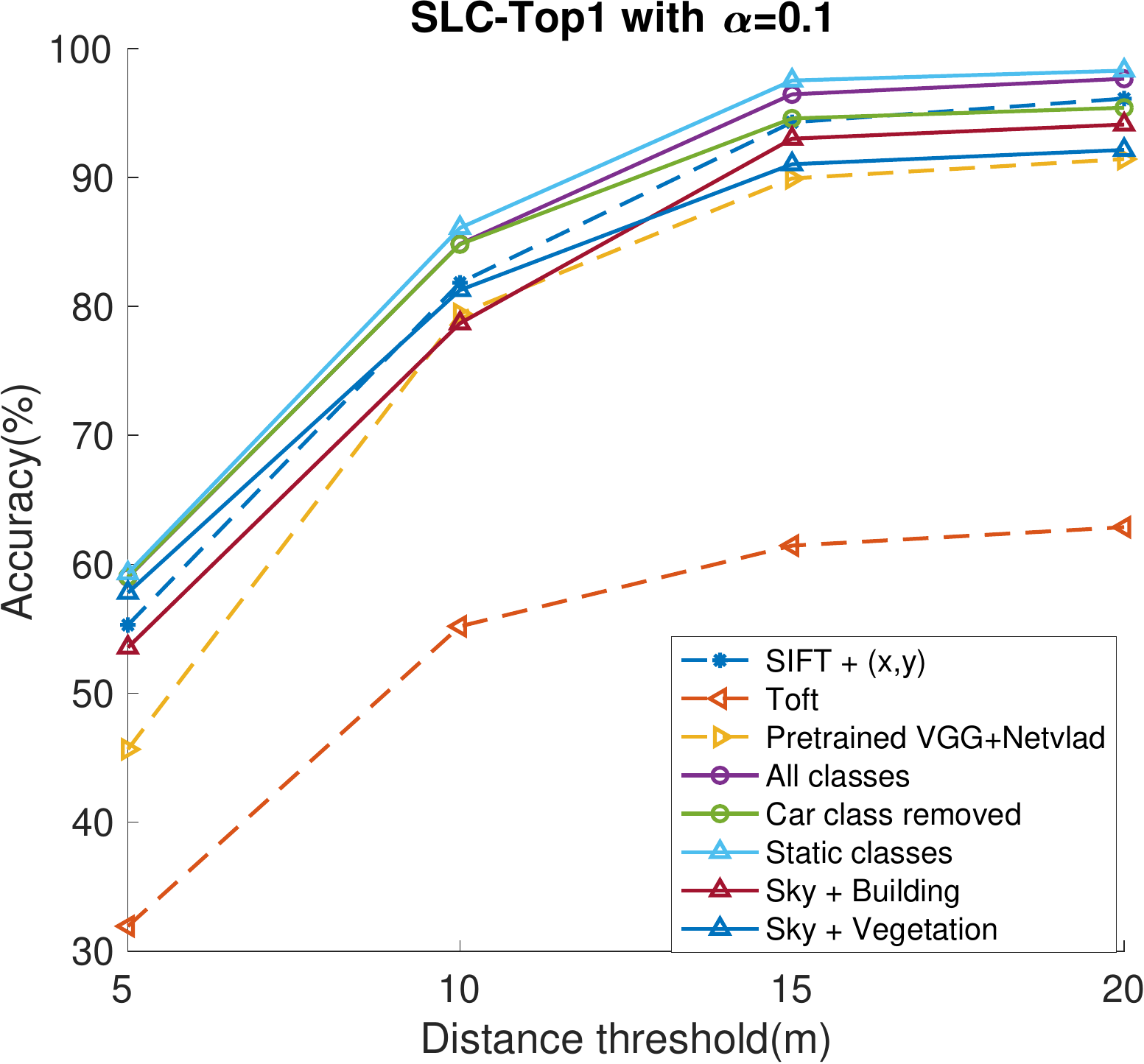} &
		\includegraphics[width=0.24\linewidth]{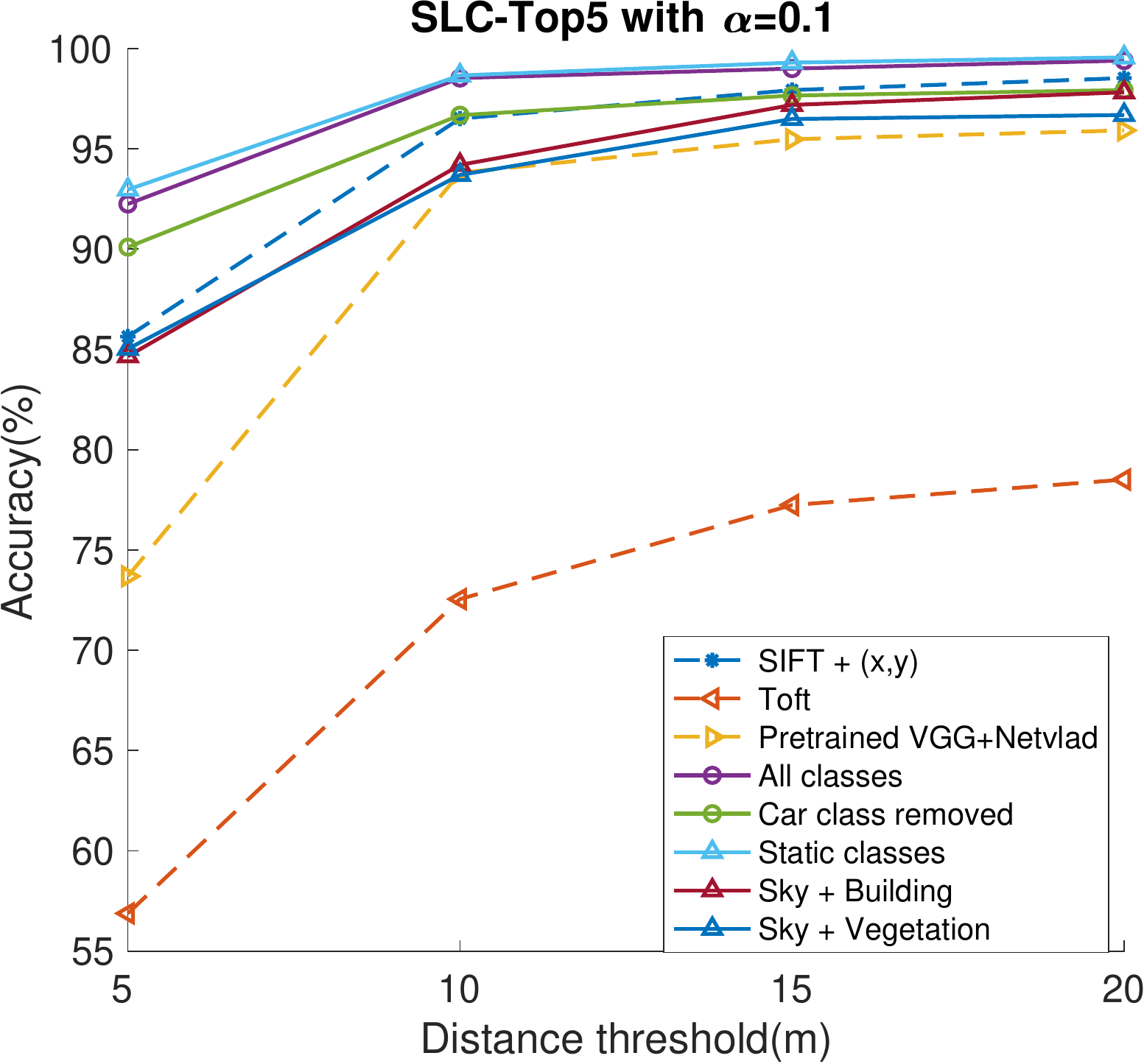} &
		\includegraphics[width=0.24\linewidth]{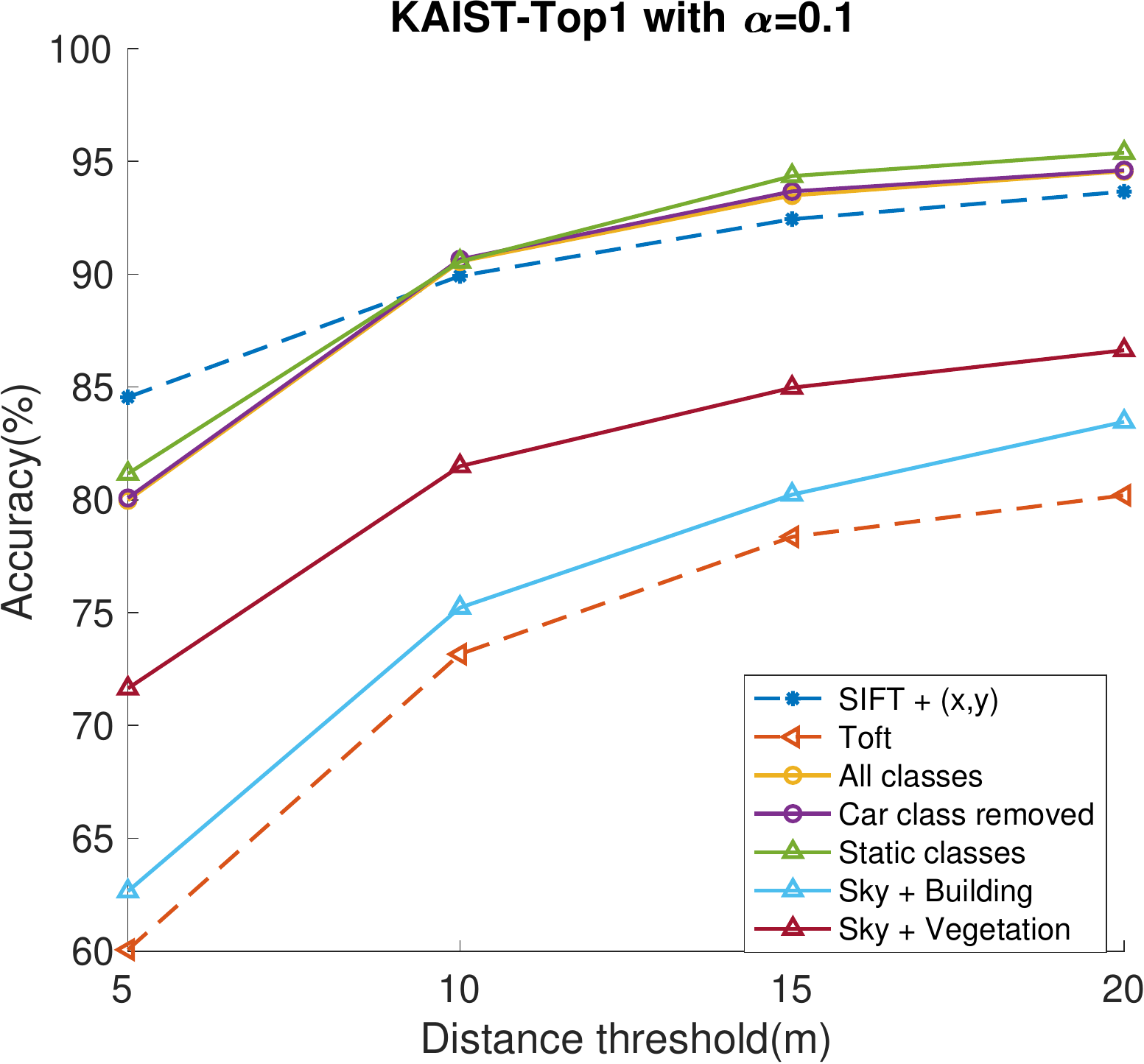} &
		\includegraphics[width=0.24\linewidth]{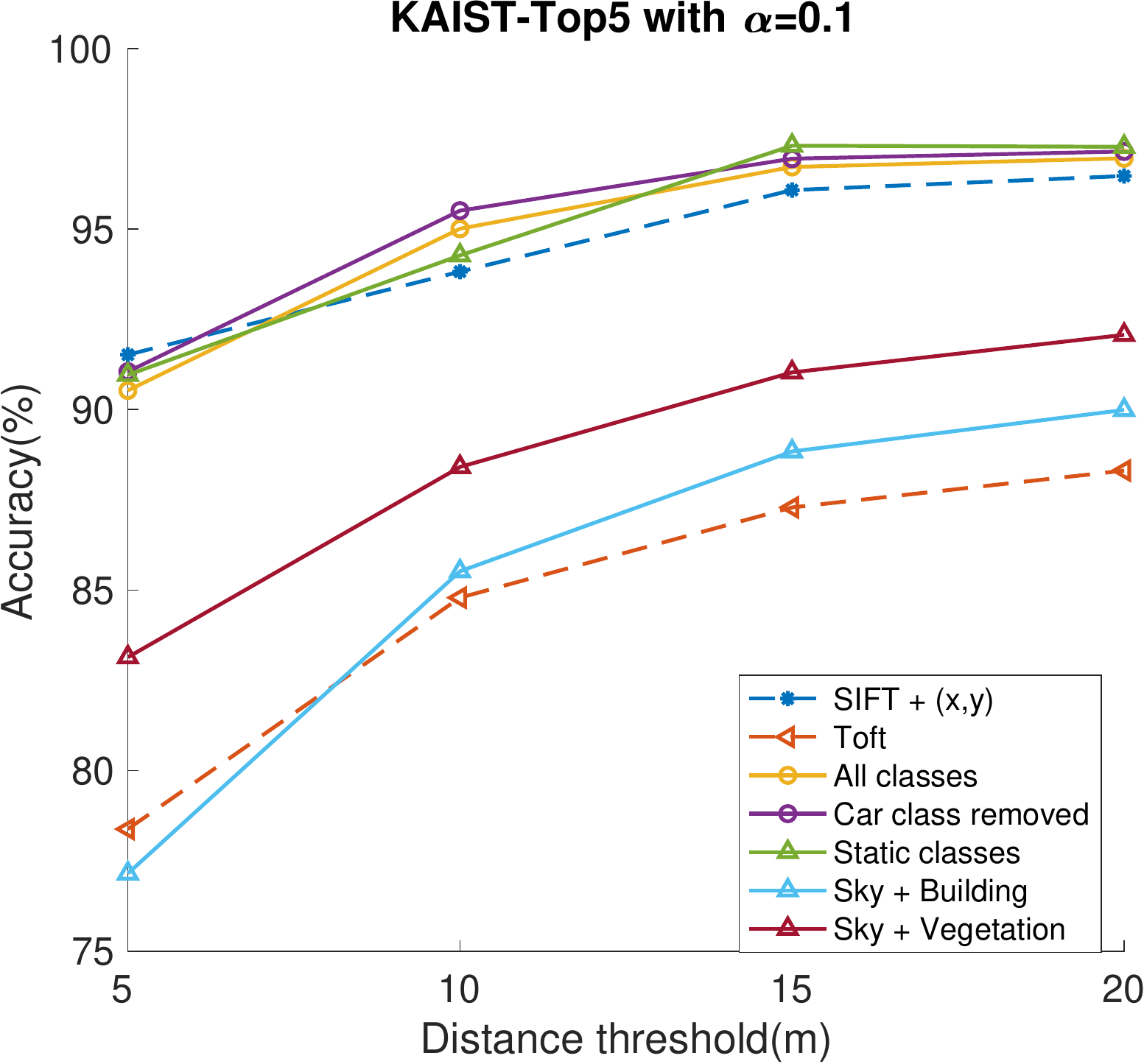}
	\end{tabular}
	\caption{Localization accuracies using weighted augmentation, with $\alpha=0.1$ found to be optimal for both SIFT and CASENet. Other settings are the same as in Figure~\ref{fig:accuracy_top1_top5}. Note Toft~\cite{toft2017long} and NetVLAD are not weighted. \label{fig:accuracy_after_weighting}}
\end{figure*}

A natural concern for such direct augmentation is the weighting of spatial coordinates compared with object class probabilities or SIFT features, which have much larger dimensions. Thus we investigate the effect of a weighted feature augmentation as $\mathbf{\bar{Y}}=[\alpha \mathbf{Y}_1, \cdots, \alpha \mathbf{Y}_K, (1-\alpha)\mathbf{Y}_x, (1-\alpha)\mathbf{Y}_y]$, where $K=19$ for CASENet and $K=128$ for SIFT, $\mathbf{Y}_x,\mathbf{Y}_y$ indicate normalized 2D spatial coordinates. In Figure~\ref{fig:alpha-scaling} and~\ref{fig:accuracy_after_weighting}, we show that combination of the two indeed achieves the best performance, and higher weights should be given to spatial coordinates due to the smaller number of dimensions.

In summary, CASENet-VLAD generally performs better than SIFT-VLAD (and also augmented SIFT-VLAD for SLC), although the augmentation sometimes makes SIFT comparable to CASENet. For example, augmented SIFT features performed better than CASENet on KAIST, since without augmentation CASENet already performed worse than SIFT (Figure~\ref{fig:accuracy_top1_top5}).
We conjectured the main reason to be the different data distributions between the KAIST and Cityscapes, leading to degraded quality of CASENet features without domain adaption. Note that another deep baseline~\cite{toft2017long}, pretrained on the Cityscapes, also performs worse than SIFT on KAIST.

%
    


\noindent \textbf{Other deep baselines}: We also compare with three deep baselines: 1) Toft~\etal's method~\cite{toft2017long}, which performs semantic segmentation using a pre-trained network~\cite{ghiasi2016laplacian} and computes a descriptor by combining histograms of static semantic classes as well as gradient histograms of building and vegetation masks in six different regions of the top half of the image; 2) VGG-NetVLAD~\cite{arandjelovic2016netvlad}; and 3) PoseNet~\cite{kendall2015posenet}, a convolutional neural network that regresses the 6-DOF camera pose from a given RGB image.
The results of the first deep baseline (our own implementation) and VGG-NetVLAD (the best pre-trained weights from the Pittsburgh dataset provided in~\cite{arandjelovic2016netvlad}) are shown to be worse than CASENet in Figure~\ref{fig:accuracy_top1_top5}.
Note for KAIST, the pretrained VGG-NetVLAD performances are very low, and even with retraining the performance is still below 30\%, thus we exclude them from Figure~\ref{fig:accuracy_top1_top5}.
For the application of PoseNet in this paper, instead of the 6-DOF output, we only regress 3 values from an image: the x-, y-location, and the orientation of the vehicle. Based on our initial experiments, we observed that the performance of PoseNet is less than 50\%. This is much lower than other methods tested in this paper (Figure~\ref{fig:accuracy_top1_top5}). We plan to investigate this further, but the high error could be due to the fact that the restricted pose parameters from the on-road vehicles (mostly straight lines and occasional turns) is insufficient to train the network.

\section{Discussion}

We proposed and validated a simple method to achieve high-accuracy localization using recently introduced semantic edge features~\cite{yu2017casenet}. While SIFT is one of the earliest feature descriptor used for localization, SIFT-VLAD is still considered as the state-of-the-art localization algorithm. We show significant improvement over the standard SIFT-VLAD, and we perform favorably to the augmented SIFT-VLAD method. While the CASENet features are trained only on cityscapes dataset, the pretrained model was sufficient for achieving state-of-the-art localization accuracy. 

Another interesting result that came out of our analysis is to show that skyline (either from building and sky, or from vegetation and sky) is a very powerful localization cue. In some of the datasets where there is too much lighting variation, the feature descriptor that just uses skylines produces results that is only marginally inferior to using all the CASENet features. 

While the main localization idea is simple, we believe that this work unifies several ideas in the community. Furthermore, it has already been shown that semantic segmentation and depth estimation are closely related to each other~\cite{Wang2015,Eigen2015}. This paper takes a step towards showing that semantic segmentation and localization are also closely related, making one more argument towards holistic scene understanding. 

In the future, we plan to consider retraining CASENet for images under bad lighting conditions. While this work was primarily about understanding how useful semantic edges are, we plan to explore more CNN-based VLAD techniques~\cite{arandjelovic2016netvlad}. We will release the SLC dataset and code for research purposes.

\addtolength{\textheight}{-0.2cm}   



%

%


\bibliographystyle{IEEEtran}
\bibliography{egbib,barcode_ref}

\end{document}